\documentclass{article}
\PassOptionsToPackage{numbers,square}{natbib}  
\usepackage[eandd,preprint]{neurips_2026}

\usepackage{fullpage}
\usepackage[numbers,square]{natbib}
\usepackage{array}
\usepackage{amssymb}
\usepackage{booktabs,tabularx,pifont}

\usepackage{amsmath,amsfonts,bm}

\def\eqref#1{equation~\ref{#1}}

\def\1{\bm{1}}

\DeclareMathAlphabet{\mathsfit}{\encodingdefault}{\sfdefault}{m}{sl}
\SetMathAlphabet{\mathsfit}{bold}{\encodingdefault}{\sfdefault}{bx}{n}

\usepackage{hyperref}
\usepackage{url}
\usepackage{booktabs}
\usepackage{graphicx}
\usepackage[table]{xcolor}
\usepackage{amssymb}
\usepackage{pifont}
\usepackage{booktabs}
\usepackage{multirow}
\usepackage{graphicx}   
\usepackage{xcolor}     
\usepackage{booktabs}
\usepackage{longtable}
\usepackage{multirow}
\usepackage[table]{xcolor}
\usepackage{booktabs}
\usepackage{pifont}
\usepackage{xcolor}
\usepackage{colortbl}
\usepackage{graphicx}
\usepackage{subcaption}
\graphicspath{{figures/}}          
\title{Meta-Video: Efficient Standardized Evaluation for Video Understanding}

\author{%
\mdseries Soumya Shamarao Jahagirdar$^{1}$ \quad Anna Kukleva$^{2}$ \quad Vishaal Udandarao$^{1}$ \quad Samuel Thomas$^{3}$ \\[2pt]
Brian Kingsbury$^{3}$ \quad Rogerio Feris$^{3,5}$ \quad James R. Glass$^{4}$ \quad Ameya Prabhu$^{6}$ \quad Hilde Kuehne$^{1,5,6}$ \\[6pt]
{\small $^{1}$University of T\"ubingen \quad $^{2}$Max Planck Institute for Informatics, SIC \quad $^{3}$IBM Research \quad $^{4}$MIT} \\[1pt]
{\small $^{5}$MIT-IBM Computing Research Lab \quad $^{6}$T\"ubingen AI Center}%
}
\date{}
\usepackage{xspace}
\newcommand{\metav}{\textbf{Meta-V}\xspace}
\newcommand{\bench}{\textbf{Meta-Video}\xspace}

\usepackage{xcolor}
\usepackage{enumitem}
\usepackage{twemojis}
\usepackage{pifont}
\newcommand{\cmark}{\ding{51}}   
\newcommand{\xmark}{\ding{55}}   
\usepackage{multirow, booktabs, graphicx, xcolor}

\usepackage{etoc}

\newlist{cfglist}{itemize}{1}
\setlist[cfglist]{leftmargin=0pt, itemindent=0pt, labelwidth=0pt, labelsep=0pt, label={}, itemsep=3pt, topsep=2pt}

\definecolor{cfgnframes}{HTML}{FFE5CC}
\definecolor{cfgsampling}{HTML}{FFD4A3}
\definecolor{cfgdecoder}{HTML}{FDBA74}
\definecolor{cfgresolution}{HTML}{F0A060}
\definecolor{cfgmodality}{HTML}{DBEAFE}
\definecolor{cfgprompt}{HTML}{BFDBFE}
\definecolor{cfgbackend}{HTML}{DCFCE7}
\definecolor{cfgbatch}{HTML}{BBF7D0}
\definecolor{cfgtemp}{HTML}{93C5FD}

\newcommand{\cfgbox}[2]{{\setlength{\fboxsep}{1.5pt}\colorbox{#1}{\textbf{#2}}}}

\newcommand{\cfgnframes}[1][]{\cfgbox{cfgnframes}{Number of frames#1}\xspace}
\newcommand{\cfgsampling}[1][]{\cfgbox{cfgsampling}{Frame sampling#1}\xspace}
\newcommand{\cfgdecoder}[1][]{\cfgbox{cfgdecoder}{Video decoder#1}\xspace}
\newcommand{\cfgresolution}[1][]{\cfgbox{cfgresolution}{Frame resolution#1}\xspace}
\newcommand{\cfgmodality}[1][]{\cfgbox{cfgmodality}{Input modality#1}\xspace}
\newcommand{\cfgprompt}[1][]{\cfgbox{cfgprompt}{Prompt#1}\xspace}
\newcommand{\cfgbackend}[1][]{\cfgbox{cfgbackend}{Inference backend#1}\xspace}
\newcommand{\cfgbatch}[1][]{\cfgbox{cfgbatch}{Batch size#1}\xspace}
\newcommand{\cfgtemp}[1][]{\cfgbox{cfgtemp}{Decoding temperature#1}\xspace}

\definecolor{cvprblue}{rgb}{0.21,0.49,0.74}
\definecolor{softgreen}{rgb}{0.23, 0.65, 0.23}
\definecolor{softred}{rgb}{0.8, 0.13, 0.13}
\definecolor{kleinblue2}{RGB}{20, 20, 125} 
\usepackage{tikz}
\usetikzlibrary{shadows}

\newcommand*\redcircled[1]{%
\tikz[baseline=(char.base)]{
  \node[shape=circle, fill=softred, draw=white, thick, drop shadow={shadow xshift=0.2ex,shadow yshift=-0.2ex}, inner sep=1pt] (char) {\textcolor{white}{#1}};
}}

\begin{document}

\maketitle
\begin{abstract}
Video understanding benchmarks are the primary yardstick for measuring progress in video-language modeling, and a growing body of work relies on them to claim improvements. The reported benchmark scores in these works, however, are almost always produced by frameworks (e.g., \texttt{lmms\_eval}) used as \textit{black boxes}: certain evaluation choices like frame sampling, preprocessing, and input formatting are inherited from preset defaults and seldom inspected carefully. We show these unexamined choices are a major source of variance. Even for the same model checkpoint, evaluation configuration alone can shift accuracy by up to $19.5$pp. 
Hence, it is unclear whether reported gains reflect genuine improvements in capability or differences in evaluation setup.
Moreover, many benchmark samples offer little evidence of visual capability, since they can be answered \textit{without} the video, and the full suites are expensive to evaluate.
We address these issues in two parts. First, we curate existing suites for discriminability and efficiency, producing \bench,  a balanced 1,000-question benchmark over 791 videos spanning diverse tasks and durations, which largely preserves model rankings, at 5$\times$ inference speedup for a full-frame count sweep experiment (4-128 frames). 
Second, we establish a standardized evaluation protocol by sweeping the configuration space to isolate choices that measurably affect results, distill them into best practices and reporting standards, and release reproducible reference scores across five models and seven benchmarks. 
We hope that the proposed evaluation framework and the \bench benchmark will support video understanding research by providing a reproducible and efficient foundation to measure progress in a reliable way.
\thanks{Project page: \url{https://soumyasj.github.io/meta-video/}}
\end{abstract}

\newcommand{\up}[1]{\textsubscript{\textcolor{green!50!black}{+#1}}}
\newcommand{\down}[1]{\textsubscript{\textcolor{red!70!black}{-#1}}}

\begin{table*}[h]
    
    \caption{\textbf{Performance reported for the Qwen2.5-VL-7B and Qwen3-VL-8B backbone.} across general video understanding and video reasoning benchmarks. Max $\Delta$ denotes the range of scores for each benchmark, differing by up to $19.5$pp despite identical model weights. Ours (reproduced) are reported for 128 frames setting.}
    \vspace{-0.15cm}
    
    \resizebox{1.1\textwidth}{!}{%
    \hspace{-2.5cm}
    \begin{tabular}{@{}lccccc@{\hspace{12pt}}ccccccc@{}}
    \toprule
    & \multicolumn{5}{c}{\textbf{General Video Understanding}} & \multicolumn{7}{c}{\textbf{Video Reasoning}} \\
    \cmidrule(lr){2-6} \cmidrule(l){7-13}
    \textbf{Model} & \textbf{VideoMME} & \textbf{TempCompass} & \textbf{MVBench} & \textbf{LongVideoBench} & \textbf{LVBench}
                   & \textbf{VideoMMMU} & \textbf{MMVU\textsubscript{mcq}} & \textbf{MMVU\textsubscript{val}} & \textbf{VSIBench} & \textbf{SciVideoBench} & \textbf{VideoMathQA} & \textbf{Minerva} \\
    \midrule
    \multicolumn{13}{@{}l}{\textit{Base model: Qwen2.5-VL-7B}} \\
    Qwen2.5-VL-7B~\citep{bai2025qwen25vl}       & 65.1 & 71.7 & 69.6 & 56.0 & 45.3 & 47.4 & -- & 50.1 & -- & -- & -- & -- \\
    \cmidrule(l){1-13}
    VideoRFT \citep{wang2025videorft}           & 52.8\down{12.3} & 69.2\down{2.5} & 59.4\down{10.2} & 56.0 & 45.3 & 47.4 & 61.3 & -- & 31.8 & -- & -- & -- \\
    Videochat-R1.5 \citep{yan2026videochat}     & 64.4\down{0.7}  & --             & 67.4\down{2.2}  & --   & --   & 46.7\down{0.7} & -- & -- & --   & -- & -- & -- \\
    Open-o3-video \citep{meng2025openo3}        & 62.4\down{2.7}  & --             & --              & --   & --   & 51.2\up{3.8}   & --   & -- & --   & -- & -- & -- \\
    Video-R2 \citep{maaz2025video}              & 66.1\up{1.0}    & 74.2\up{2.5}   & 68.1\down{1.5}  & 61.3\up{5.3} & -- & 50.3\up{2.9} & 67.4 & -- & 37.7 & 16.4 & 26.7 & 35.3 \\
    Viss-R1 \citep{fang2026viss}                & 56.6\down{8.5}  & 72.6\up{0.9}   & 59.0\down{10.6} & --   & --   & 48.1\up{0.7}   & 60.0 & -- & 30.1 & -- & -- & -- \\
    Rethinking CoT \citep{zhong2025rethinking}  & 55.5\down{9.6}  & --             & 63.2\down{6.4}  & 52.5\down{3.5} & 34.2\down{11.1} & -- & 63.4 & -- & -- & -- & -- & -- \\
    STORM \citep{liang2026storm}                & 55.4\down{9.7}  & 69.9\down{1.8} & 60.5\down{9.1}  & --   & --   & --             & --   & -- & --   & -- & -- & -- \\
    CoVER \citep{wang2026see}                   & 63.1\down{2.0}  & --             & --              & 56.0 & 42.0\down{3.3} & -- & --   & -- & --   & -- & -- & -- \\
    Video-VER \citep{luo2026thinking}           & 59.2\down{5.9}  & 73.7\up{2.0}   & 63.6\down{6.0}  & --   & --   & 47.3\down{0.1} & 64.2 & -- & 32.3 & -- & -- & -- \\
    Video-TwG \citep{chen2026think}             & 46.6\down{18.5} & --             & --              & --   & --   & 44.4\down{3.0} & --   & -- & --   & -- & -- & -- \\
    SDRL \citep{wang2026reinforcing}            & 56.5\down{8.6}  & 72.5\up{0.8}   & 63.3\down{6.3}  & --   & --   & 48.4\up{1.0}   & 65.4 & -- & 29.1 & -- & -- & -- \\
    Video-ToC \citep{tan2026video}              & 59.6\down{5.5}  & 72.9\up{1.2}   & 59.2\down{10.4} & --   & --   & 50.4\up{3.0}   & 60.0 & -- & 31.4 & -- & -- & -- \\
    VidBridge-R1 \cite{chen2026vidbridge}       & 59.4\down{5.7}  & --             & 58.7\down{10.9} & 50.6\down{5.4} & -- & --           & --   & -- & --   & -- & -- & -- \\
    TW-GRPO~\cite{dang2026reinforcing}          & 56.5\down{8.6}  & 72.5\up{0.8}   & 63.3\down{6.3}  & --   & --   & --             & 65.4   & -- & --   & -- & -- & -- \\
    VideoAuto-R1~\cite{liu2026videoauto}        & 66.0\up{0.9}    & --             & 67.1\down{2.5}  & 60.9\up{4.9} & -- & 52.7\up{5.3}   & --   & -- & --   & -- & -- & -- \\
    MOSS-ChatV~\cite{tao2025moss}               & 59.7\down{5.4}  & 72.2\up{0.5}   & 67.1\down{2.5}  & --   & --   & 48.1\up{0.7}   & --   & -- & --   & -- & -- & -- \\
    STRIVE~\cite{bahrami2026strive}             & 57.8\down{7.3}  & 72.8\up{1.1}   & --              & --   & --   & 48.6\up{1.2}   & --   & -- & 34.3 & -- & -- & -- \\
    RLER~\cite{yang2026reinforce}               & 65.1            & 69.2\down{2.5} & 67.5\down{2.1}  & 56.0 & 42.0\down{3.3} & 47.4 & --   & -- & 37.4 & -- & -- & -- \\
    VideoKR~\cite{fu2026videokr}                & 65.1            & --             & 66.3\down{3.3}  & 60.9\up{4.9} & -- & 51.1\up{3.7}   & -- & 55.7 & --   & 28.1 & -- & -- \\
    AdaThinkV~\cite{tian2026adathinkv}          & 66.0\up{0.9}    & --             & 67.1\down{2.5}  & 60.9\up{4.9} & -- & 54.7\up{7.3}   & 66.2 & -- & --   & 27.9 & 29.8 & -- \\
    \cmidrule(l){1-13}
    \textit{Max $\Delta$}                       & 19.5 & 5.0 & 10.9 & 10.7 & 11.1 & 10.3 & 7.4 & 5.6 & 8.6 & 11.7 & 3.1 & 0.0 \\
    
    \rowcolor{blue!8}
    \textbf{Ours (Reproduced)}  & 65.6\up{0.5} & 73.4\up{1.7} & 68.0\down{1.6} & 60.2\up{4.2} & 42.1\down{3.2} & 38.8\down{8.6} &  64.6 & 51.9\up{1.8} & 38.3 & 27.9 & 26.9 & 35.3  \\
    
    \midrule
    
    \multicolumn{13}{@{}l}{\textit{Base model: Qwen3-VL-8B-Instruct}} \\
    Qwen3-VL-8B-Instruct~\citep{bai2025qwen3}   & 71.4 & --   & 68.7 & --   & 58.0 & 65.3 & -- & 58.7 & -- & -- & -- & -- \\
    
    \cmidrule(l){1-13}
    LLaVA-OneVision-2 \citep{an2026llavaov2}    & 71.4            & 74.3 & 69.0\up{0.3}   & 68.0 & 58.0            & --             & --   & -- & -- & --   & --   & -- \\
    VSeek \citep{goel2026incentivizing}         & 63.7\down{7.7}  & --   & --             & 54.5 & 38.7\down{19.3} & --             & --   & -- & -- & --   & --   & -- \\
    OneThinker \citep{feng2026onethinker}       & 63.3\down{8.1}  & --   & --             & 61.7 & --              & 63.3\down{2.0} & --   & -- & -- & --   & --   & -- \\
    VideoKR \citep{fu2026videokr}               & 68.2\down{3.2}  & --   & 67.9\down{0.8} & 61.6 & --              & 61.8\down{3.5} & -- & 59.6 & -- & 33.4 & --   & -- \\
    InternVideo3 \citep{yan2026internvideo3}    & 71.4            & --   & --             & 62.4 & 58.0            & --             & --   & -- & -- & --   & --   & -- \\
    AdaThinkV \citep{tian2026adathinkv}         & 72.5\up{1.1}    & --   & 69.4\up{0.7}   & 67.6 & --              & 61.0\down{4.3} & 69.9 & -- & -- & 33.1 & 30.7 & -- \\
    \cmidrule(l){1-13}
    \textit{Max $\Delta$}                       & 9.2 & -- & 1.5 & 13.5 & 19.3 & 4.3 & -- & 0.9 & -- & 0.3 & -- & -- \\
    
    \rowcolor{blue!8}
    \textbf{Ours (Reproduced)} & 69.2\down{2.2} & 74.2 & 69.2\up{0.5} & 63.6 & 46.5\down{11.5} & 56.6\down{8.7} & 69.1 & 58.7 & 59.7 & 34.5 & 31.6 & 36.0 \\
    \bottomrule
    \end{tabular}
    }
    \label{tab:table1_performance_discrepancy}
\end{table*}

\section{Introduction}
\label{introduction}
Benchmarks measure progress in video understanding, with researchers often using benchmark scores to claim improvements over prior methods or base models. Such claims assume that scores for the same checkpoint on the same benchmark are fairly comparable across evaluations and papers. We find that they are not. Table~\ref{tab:table1_performance_discrepancy} compares scores reported in recent papers for two widely used checkpoints, Qwen2.5-VL-7B~\citep{bai2025qwen25vl} and Qwen3-VL-8B-Instruct~\citep{bai2025qwen3}, across 11 video benchmarks. Several methods report gains over the Qwen2.5-VL-7B baseline that fall below our reproduction of the same baseline. When we evaluate these methods under identical settings, they don't improve consistently over the baseline across varying frame counts. We see that reported performance varies substantially across papers by up to 19.5pp on VideoMME on the same checkpoint.  
These differences far exceed the 2–5pp gains commonly showcased by these same papers over the base model. This makes it hard to tell whether an apparent gain claimed in these prior works, stems from a real capability improvement in the model, a difference in evaluation configuration, or both.

This shows that video evaluation can be especially sensitive to eval configuration. For example, \cfgsampling determines which events are visible to a model. 
Which \cfgdecoder is used, how \cfgnframes and \cfgresolution are set determine  the detail and temporal structure visible to the model. 
Further, the \cfgmodality (input a video file or sequence of frames) and text \cfgprompt can also affect performance.
Yet, papers use evaluation frameworks like \texttt{lmms\_eval} as \textit{black boxes}, leaving such details underspecified. Additionally, the quality of benchmark samples can affect the value and cost of evaluation. Some benchmark samples can be answered without even seeing the video.
Further, because model rankings can differ across benchmarks, robust comparisons demand aggregating evaluations on multiple datasets, with very large inference costs. 
As such, a reliable evaluation supporting fast model development calls for \redcircled{1} a clean set of benchmark samples, \redcircled{2} low evaluation cost while preserving ranking, and \redcircled{3} a standardized evaluation protocol.

To address \redcircled{1} and \redcircled{2}, we introduce \bench (Sec.~\ref{subsec:meta-videobench}), a compact video understanding benchmark of 1,000 questions over 791 videos, balanced across source datasets and spanning diverse tasks and durations. We construct it by filtering out low-quality samples to further improve evaluation fidelity and computational efficiency \citep{joshi2026datbench, ghosh2025onebench}. Specifically, we remove questions solvable without the video, since these cannot measure video understanding. \bench largely preserves model rankings over its source benchmarks, separates models more sharply, and requires $5\times$ less inference compute than the reference benchmarks for a full sweep over frame counts, which is the most expensive evaluation setting (Table~\ref{tab:table2_allparameters}(b)). 
To tackle \redcircled{3}, we identify key parameters in video evaluation pipelines and systematically measure their effects on performance (Sec.~\ref{method}). 
We find that input representation and inference backend substantially change scores. 
In particular, switching between vLLM and Hugging Face while keeping their respective default settings can change performance by more than 5pp, with gains decreasing once the defaults are fixed.
Further, we show that extracted \texttt{.jpg} frames yield higher scores when few frames are sampled, while \texttt{.mp4} input performs better with larger frame budgets.
We observe that how many frames and which frames (frame sampling) the model sees is one of the largest, most consistent factors influencing  performance.
Our systematic evaluation identifies the major impact factors and proposes a concise way to compile them in the form of \textit{evaluation cards} to allow for a better, transparent, and reproducible practice in video understanding benchmarking.


Together, our contributions are:
(1) We introduce \bench, a compact, balanced benchmark that preserves ranking trends across its source benchmarks (reliability), uses one-fifth of the inference compute for a full frame-count sweep (efficiency), and makes differences between models clearer (discriminability).
(2) We examine the current state of video understanding evaluation and show that variation in often unspecified evaluation protocols can explain substantial discrepancies in reported scores.
(3) We release a revised evaluation suite and an extensive set of reference configurations to improve reproducibility and support future comparisons.
We hope our findings and  standardized evaluation protocols offer a transparent and less costly way to evaluate video understanding, in a reliable manner.


\section{Related work}
\label{related_work}
\noindent\textbf{Video Understanding Models.} General purpose vision-language models have the capability to ingest multi-frame input and hence can perform video understanding tasks. Models such as Qwen2.5-VL \citep{bai2025qwen25vl}, InternVL \citep{zhu2025internvl3}, LLaVA-OneVision \citep{li2024llavaov,an2025llava15onevision,an2026llavaov2}, PerceptionLM \citep{cho2026perceptionlm}, Molmo \citep{deitke2025molmo}, LLaVA-Next-Interleave \citep{li2024llava} etc., are widely used as standard open-weight VLMs. Other works enable such models to interpret and reason over dynamic visual content \citep{shu2025audio,cheng2024videollama,yan2024visa,chen2024videollm,li2024llama,zhang2024long,maaz2024video,maaz2024videogpt+,yang2025qwen3,zhu2024minigpt}. More recently, video reasoning models have surfaced with the ability to reason over temporal and causal relationships across sequences of frames
~\citep{meng2025videocap,wang2025timer1,sun2025videosalmon,zhang2025tinyllava,wang2025videorft}.


\noindent\textbf{Video Understanding Benchmarks.}
Multiple benchmarks have been proposed to evaluate video understanding e.g., VideoMME~\citep{fu2025videomme}, LongVideoBench~\citep{wu2024longvideobench}, MVBench~\citep{li2024mvbench}, LVBench~\citep{wang2025lvbench} etc. Further, several benchmarks target reasoning, such as, VideoMMMU~\citep{hu2025videommmu}, VideoMathQA~\citep{rasheed2026videomathqa}, SciVideoBench~\citep{deng2025scivideobench}, VideoReasonBench~\citep{liu2026videoreasonbench}, VCRBench~\citep{qi2025vcr}, TempCompass~\citep{liu2024tempcompass}, MMVU~\citep{zhao2025mmvu}, MLVU~\citep{zhou2025mlvu}, Minerva~\citep{nagrani2025minerva}, EgoSchema~\citep{mangalam2023egoschema} etc. However, these benchmarks have key limitations~\citep{udandarao2025solving,yang2026cambrian}. For e.g., many questions can be answered \textit{without} any visual information~\citep{fu2026videokr}, as we show in Sec.~\ref{appendix}. In addition, because of their large size, evaluation is extremely time-consuming, discouraging rapid iteration for fast model development.

\noindent\textbf{Validity of Benchmark Evaluations.}
Recent works have tried to improve the validity of benchmarks. Answer-Matching~\citep{chandak2025answer} reformulates multiple-choice tasks into a generative setting, MMBench~\citep{liu2024mmbench} uses circular evaluation techniques, VMCBench automatically scales up question generation process~\citep{zhang2025automated}, OneBench proposes aggregating heteregeneous measurements across benchmarks~\citep{ghosh2025onebench}, and DataComp-VLM~\citep{farina2026datacomp} uses stability-monotonicity filtering. DatBench~\citep{joshi2026datbench} proposed a cleaned evaluation suite that expedites evaluation and maintains discriminative power. Sober Reasoning~\citep{hochlehnert2025sober} provides an empirical evaluation of mathematical reasoning benchmarks showcasing sensitivity to various implementation choices like random seed, prompt formatting, and hardware configurations. 
All these prior works focus mainly on language and image domains, with the video domain taking a back seat.
Further, evaluation repositories such as \texttt{lmms\_eval}~\citep{zhang2025lmms} and \texttt{VLMEvalKit}~\citep{duan2024vlmevalkit} provide a framework to evaluate various multimodal benchmarks with multiple models. However, they are often used as black boxes, which makes results hard to compare.
We close this gap with a standardized evaluation protocol for video and a new benchmark, \bench, built for discriminability and fast iteration.




\section{\bench: An Efficient and Discriminative Benchmark}
\label{subsec:meta-videobench}

\begin{figure*}[t]
  \centering
  \begin{subfigure}[t]{0.45\textwidth}
    \centering
    \includegraphics[width=\linewidth]{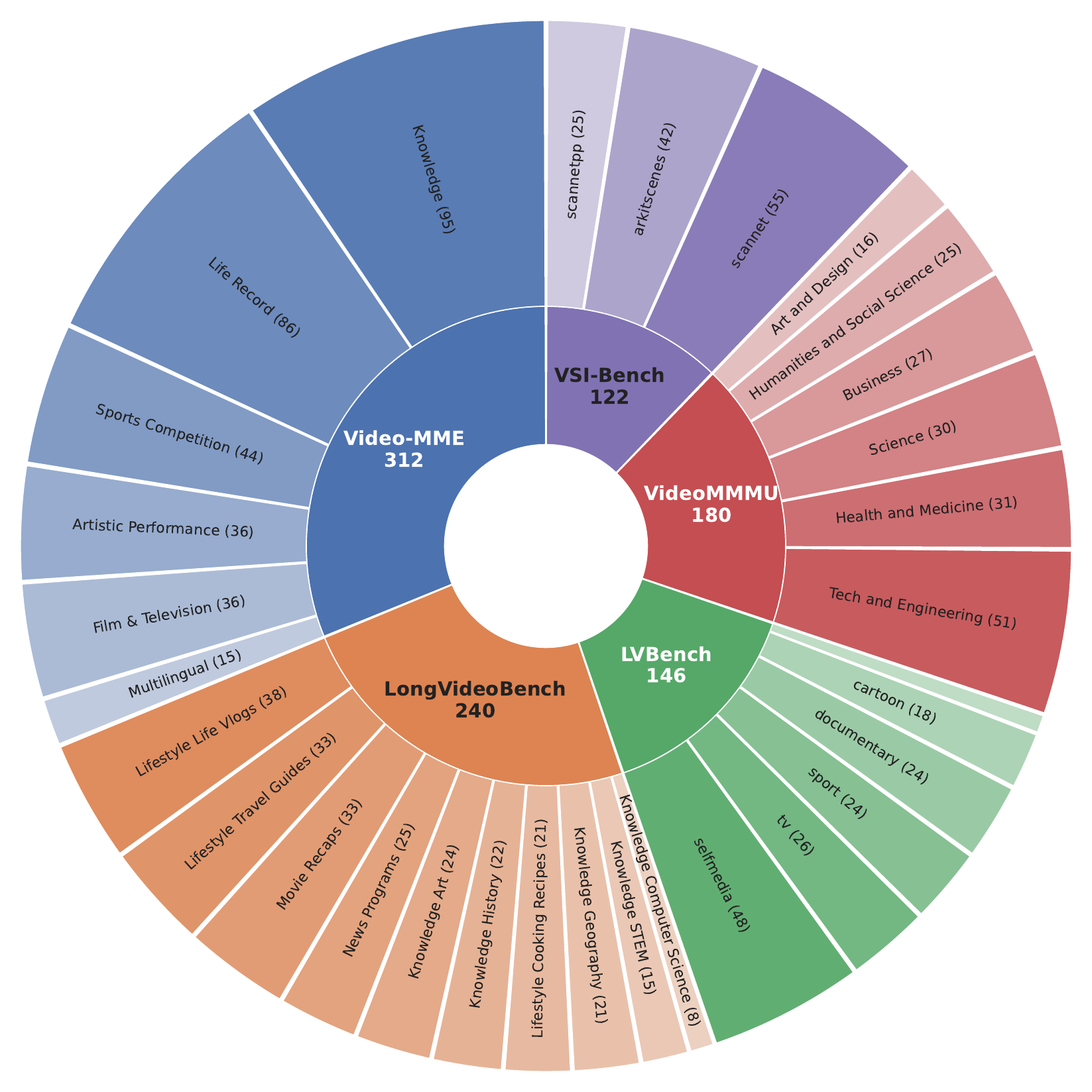}
    \caption{Source benchmarks and their categories.}
    \label{fig:stats-domains}
  \end{subfigure}%
  \hfill
  \begin{subfigure}[t]{0.45\textwidth}
    \centering
    \includegraphics[width=\linewidth]{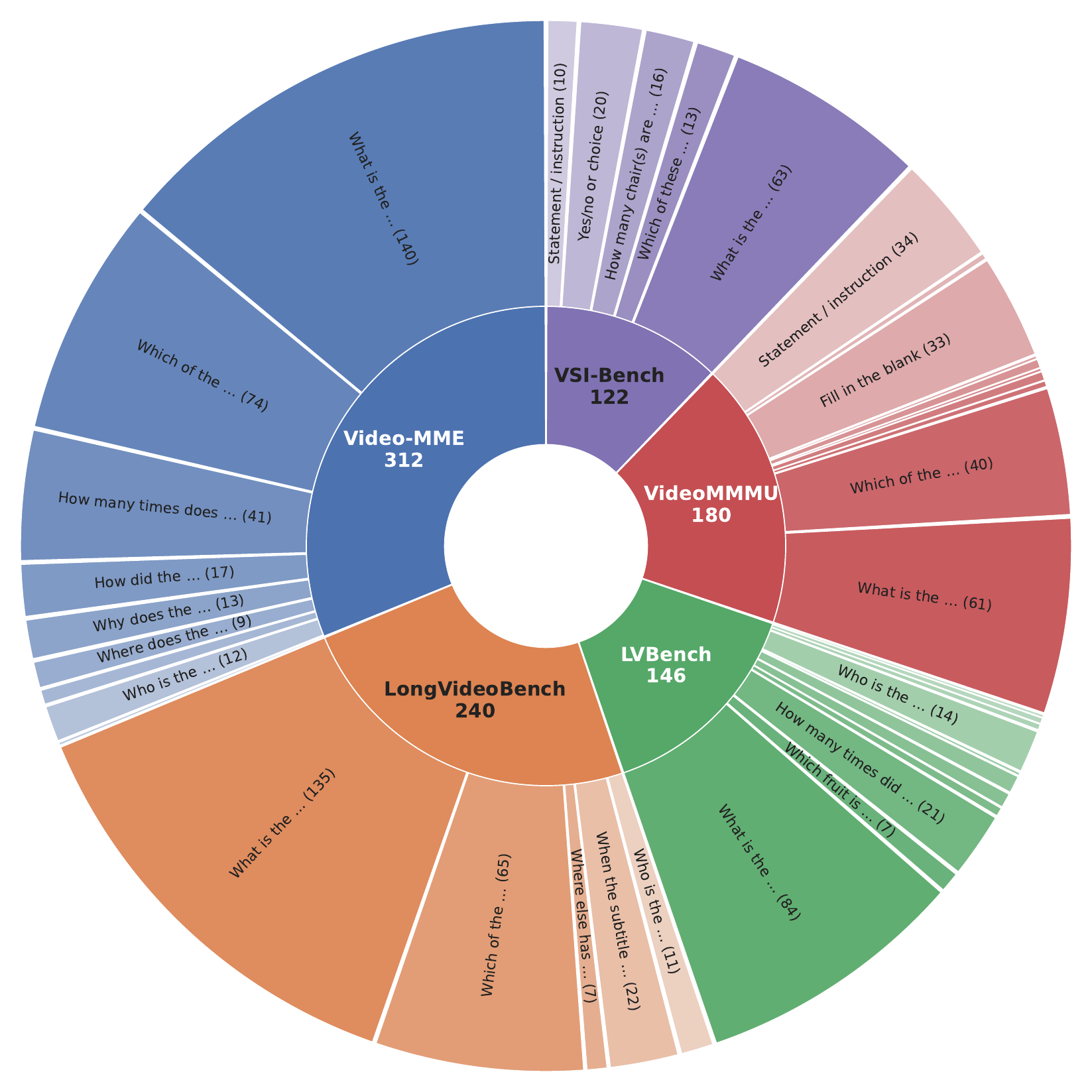}
    \caption{Question types within each source benchmark. 
    }
    \label{fig:stats-qtypes}
  \end{subfigure}
  \begin{subfigure}[t]{0.45\textwidth}
    \centering
    \includegraphics[width=\linewidth]{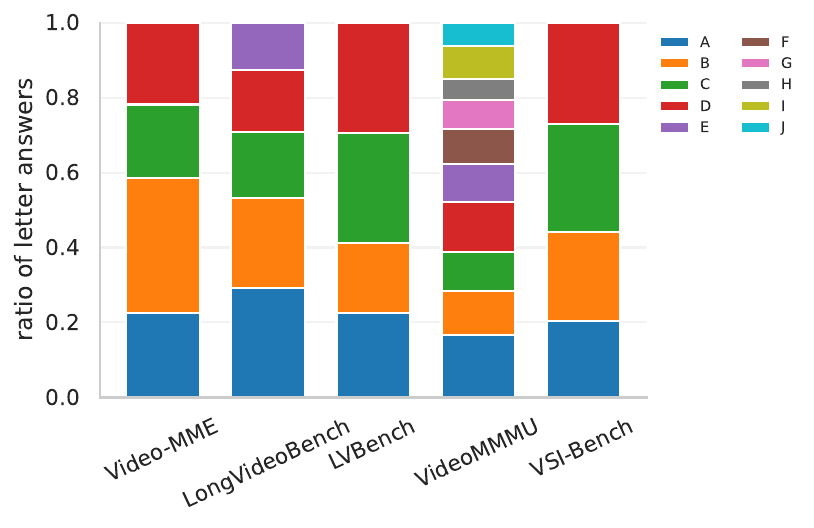}
    \caption{Distribution of correct answer letters per benchmark (63 non-letter answers excluded).}
    \label{fig:stats-letters}
  \end{subfigure}%
  \hfill
  \begin{subfigure}[t]{0.45\textwidth}
    \centering
    \includegraphics[width=\linewidth]{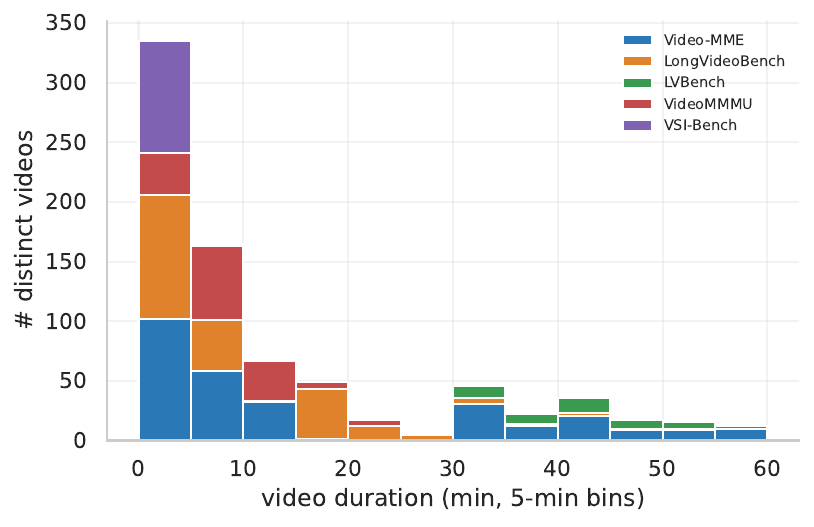}
    \caption{Video duration distribution on \bench.}
    \label{fig:stats-durations}
  \end{subfigure}
  \vspace{-2mm}
  \caption{\textbf{\bench statistics.} The benchmark contains 1{,}000 QA pairs over 791 distinct videos.}
  \label{fig:metav-benchmark-stats}
  \vspace{-4mm}
\end{figure*}

Evaluating video models across multiple benchmarks and configurations to ensure robustness is expensive, particularly for researchers with limited compute. Reliable comparisons also require questions that test video understanding: questions that models can answer without visual input can inflate scores and obscure differences between models. To address these problems, we introduce \bench, a  compact benchmark of videos and question-answer pairs selected from existing datasets, 
while reducing inference cost.

\subsection{Source Benchmarks and Construction}
We select video QA pairs from the evaluation sets of Video-MME, LongVideoBench, LVBench, and VideoMMMU, and VSI-Bench (subset of 200 QA pairs), which cover a range of video types, durations, and question categories. To filter questions that may be answerable without visual input, we replace all frames with black frames and evaluate Qwen2.5-VL-7B-Instruct, Qwen3-VL-8B-Instruct, and InternVL3.5-8B. We discard each question that at least two or more models answer correctly, then sample QA pairs from the remaining questions in each source benchmark (see Appendix~\ref{appendix} for details).

\subsection{Statistics}
\label{subsec:metav_stats}
%
\noindent\textbf{Question types.} \bench contains 1{,}000 QA pairs from 791 distinct videos, spanning multiple domains and question types, as shown in Figures~\ref{fig:stats-domains} and~\ref{fig:stats-qtypes}. Of these, 937 are multiple-choice questions and 63 require numerical free-form answers. Figure~\ref{fig:stats-letters} shows the distribution of correct answer options, with an even distribution across options. The benchmark includes 564 questions with four options, 185 with five, 170 with more than five, of which 157 questions have exactly ten options, and  18 questions with fewer than four options. Random guessing on multiple-choice questions and answering $0$ on numerical questions yields a score of $20.3$. Always selecting option B, the most frequent correct option, and answering $0$ on numerical questions yields $23.3$.


\noindent\textbf{Video duration.} \bench spans video durations up to one hour, as illustrated in Figure~\ref{fig:stats-durations}. The duration groups comprise 335 short videos ($<5$ minutes), 281 medium videos ($5$--$20$ minutes), and 175 long videos ($20$--$60$ minutes). This range matters because benchmarks described as testing long-video understanding differ substantially in duration: videos average 41.1 minutes in the Video-MME long split, 7.9 minutes in LongVideoBench, and 67.3 minutes in LVBench. Appendix~\ref{appendix} provides duration statistics for each source benchmark.

\textbf{Results.} 
Table~\ref{tab:table2_allparameters} evaluates models under different parameter settings and compares \bench with the average across all the benchmarks. \bench follows the trend of the average of multiple benchmarks, highlighting the fact that using \bench can show the trend for any new model to be tested, with the additional advantage of being very efficient compared to running all benchmarks at all settings, which is computationally expensive. 

\section{Standardizing Evaluations for Video Understanding}\label{method}


Evaluation choices can change both model scores and rankings. We examine these choices at multiple stages: inference and decoding (Sec.~\ref{inference_and_decoding}), frame extraction (Sec.~\ref{frame_extraction}), model input (Sec.~\ref{model_input}), and prompt design (Sec.~\ref{promptdesign}). Based on this, we fix the most influential factors to obtain a reproducible evaluation protocol (Sec.~\ref{protocol}), while documenting detailed configurations in evaluation cards. Finally, we provide comprehensive evaluations across seven benchmarks and five models (Sec.\ref{subsec:eval_models}).

\begin{figure*}[t]
	\centering
	\includegraphics[width=.99\textwidth]{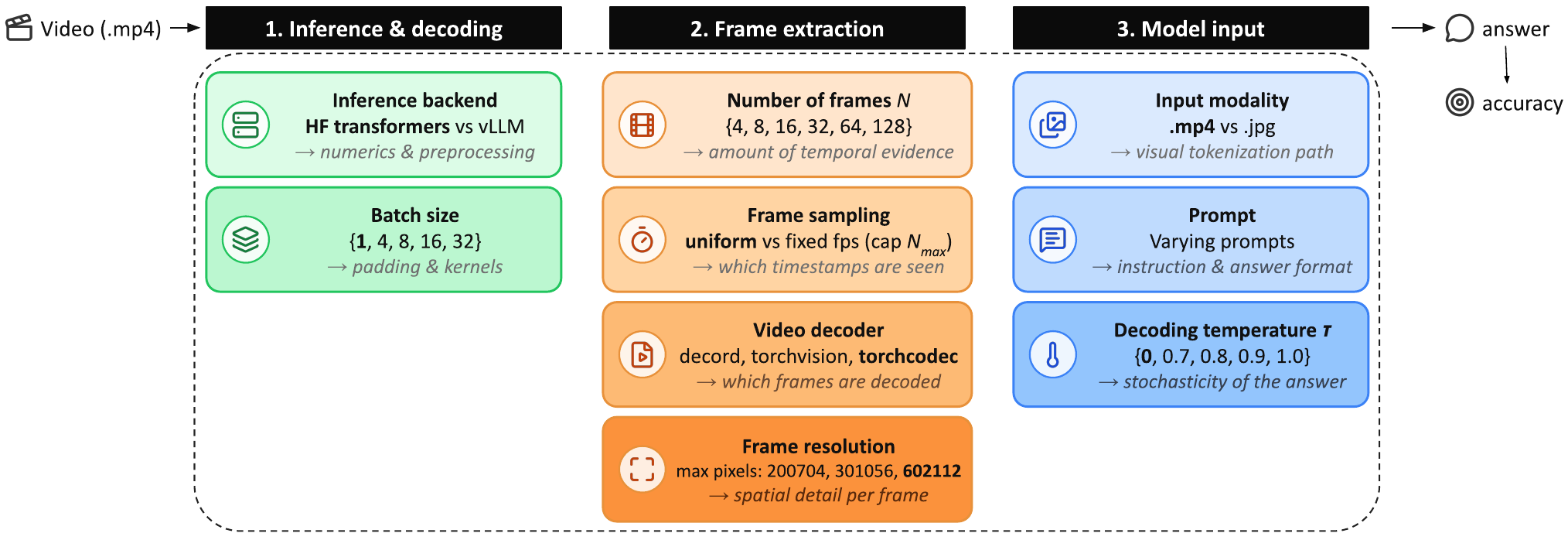}
    \vspace{-2mm}
    \caption{\textbf{Evaluation framework.} Parameters in the evaluation pipeline that affect reported accuracy for video understanding benchmarks. 
    }
	\label{fig:protocol_figure}
    \vspace{-6mm}
\end{figure*}



\subsection{Inference and decoding}
\label{inference_and_decoding}

\cfgbackend[.] Video question-answering frameworks such as \texttt{lmms\_eval} support multiple inference backends, including the reference Hugging Face (HF) implementation~\citep{huggingface} and optimized engines such as vLLM~\citep{kwon2023efficientvllm}. These backends can differ in both numerical computation and preprocessing. In our experiments, HF produces identical scores across repeated runs with fixed seeds, whereas vLLM exhibits significant variations even with fixed seeds, potentially due to batching.


\cfgbatch[.] Batch composition and size can affect padding and kernel selection, changing outputs even under greedy decoding. Although batch size has little effect on aggregate performance in our experiments, it remains a source of variation and should be reported.
%
%
%
Table~\ref{tab:table2_allparameters}(a) compares the impact of inference backends and batch sizes. Scores within each framework are stable, with a mean standard deviation of $0.15$ percentage points, but scores between backends differ by $4$--$6$ points on TempCompass, MVBench, and \bench. On \bench, for example, the same model scores approximately $24\%$ with HF and $30\%$ with vLLM. We find that this discrepancy is attributable to the evaluated vLLM implementation converting video inputs into individual images, bypassing the Qwen models' spatiotemporal pooling. We provide a fix lmms\_eval\_video implementation that restores this pooling and aligns vLLM preprocessing with HF. Appendix~\ref{appendix} describes the fix and its results.

\newcolumntype{Y}{>{\raggedright\arraybackslash}p{1.3cm}}  
\newcolumntype{Z}{>{\raggedright\arraybackslash}p{2.74cm}} 
\newcolumntype{A}{>{\centering\arraybackslash}p{2.5cm}}    
\newcolumntype{M}{>{\columncolor{blue!8}\centering\arraybackslash}p{2.5cm}} 
\newcommand{\SubWidth}{\linewidth}
\newcommand{\BenchHead}{Video-MME & TempComp & MVBench & LongVideoB & Avg. & \textbf{\textcolor{blue}{Meta-V}} \\}
\newcommand{\SubHeadII}[2]{#1 & #2 & \BenchHead}
\newcommand{\SubHeadI}[1]{#1 & \BenchHead}
\begin{table*}[t]
\centering
\caption{\textbf{Sensitivity of evaluation settings and parameters on Qwen2.5-VL-7B.}
Each sub-table varies one setting; all other settings are fixed as listed.
All runs use the official \texttt{lmms\_eval} prompt.
All values are reported as accuracy (\%), except \textit{Runtime}, which gives the total runtime in minutes of all runs in the block above them.
Avg.\ is the mean accuracy over the four benchmarks; in the \textit{Runtime} rows, Avg.\ gives the summed runtime over all benchmarks.
$\pm$ denotes the standard deviation over repeated runs.
Column labels: nf (number of frames), bck (backend), bs (batch size), rdr (video reader), px (max pixels), samp (sampling; $\leq n$ means 1\,fps capped at $n$ frames), $T$ (temperature).}
\begin{minipage}[t]{0.48\textwidth} 
\centering
{\footnotesize(a)\cfgbackend \texttwemoji{2b50} and \cfgbatch }\\
\parbox{\linewidth}{\centering\tiny\textit{Fixed:} 602k px, uniform sampling, torchcodec, \texttt{.mp4}, $T{=}0$, 4 frames 
}\\[3pt]

\resizebox{\SubWidth}{!}{%
\begin{tabular}{ll cccc | c | c}
\toprule
\SubHeadII{bck}{bs}
\midrule
HF & 1  & 50.0 & 64.4 & 59.2 & 50.3 & 56.0 & \textbf{24.8} \\
 & 4  & 50.0 & 64.7 & 59.6 & 50.5 & 56.2 & {24.4} \\
 & 8  & 50.0 & 65.0 & 59.3 & 50.7 & 56.2 & {24.1} \\
 & 16 & 50.1 & 64.8 & 59.2 & 50.4 & 56.1 & {24.6} \\
 & 32 & 50.0 & 64.9 & 59.3 & 50.4 & 56.1 & {24.6} \\
\rowcolor{gray!15}\textit{Runtime} & & \textit{157} & \textit{46.6} & \textit{166} & \textit{77.2}  & \textit{447} & {133} \\
\midrule
vLLM & 1  & 50.4{\tiny$\pm$0.1} &  69.{\tiny$\pm$0.1} & 64.4{\tiny$\pm$0.1} & 52.1{\tiny$\pm$0.1} & 58.9 & \textbf{31.0{\tiny$\pm$0.1}}\\
 & 4  & 50.3{\tiny$\pm$0.0} & 69.2{\tiny$\pm$0.2} & 64.4{\tiny$\pm$0.0} & 51.9{\tiny$\pm$0.2} & 59.0 & {30.5{\tiny$\pm$0.2}} \\
 & 8  & 50.2{\tiny$\pm$0.2} & 69.2{\tiny$\pm$0.2} & 64.3{\tiny$\pm$0.2} & 51.9{\tiny$\pm$0.1} & 58.9 & {30.6{\tiny$\pm$0.5}} \\
 & 16 & 50.3{\tiny$\pm$0.1} & 69.2{\tiny$\pm$0.4} & 64.4{\tiny$\pm$0.2} & 51.7{\tiny$\pm$0.0} & 58.9 & {30.4{\tiny$\pm$0.3}} \\
 & 32 & 50.3{\tiny$\pm$0.1} & 69.1{\tiny$\pm$0.1} & 64.3{\tiny$\pm$0.1} & 51.8{\tiny$\pm$0.1} & 58.9 & {30.4{\tiny$\pm$0.1}} \\
\rowcolor{gray!15}\textit{Runtime} & & \textit{155} & \textit{55.6} & \textit{257} &  \textit{84} & \textit{551} & {112} \\
\bottomrule
\end{tabular}}
\end{minipage}
\hfill
\begin{minipage}[t]{0.48\textwidth} 
\centering
{\footnotesize(b)\cfgmodality \texttwemoji{2b50} and \cfgnframes} \texttwemoji{2b50}\\
\parbox{\linewidth}{\centering\tiny\textit{Fixed:} 602k px, uniform sampling, HF, batch 1, $T{=}0$; \texttt{.mp4}; torchcodec}\\[3pt]
\resizebox{\SubWidth}{!}{%
\begin{tabular}{ll cccc | c | c}
\toprule
\SubHeadII{in}{nf}
\midrule
\texttt{.mp4} & 4   & 50.0 & 64.4 & 59.2 & 50.3 & 56.0 & {24.8} \\
 & 8   & 53.3 & 70.3 & 63.4 & 52.7 & 59.9 & {28.2} \\
 & 16  & 58.0 & 71.8 & 66.7 & 56.1 & 63.1 & {33.4} \\
 & 32  & 61.4 & 73.3 & 68.5 & 58.7 & 65.5 & {36.1} \\
 & 64  & 63.5 & 74.1 & 69.0 & 60.2 & 66.7 & {39.2} \\
 & 128 & 65.5 & 73.4 & 68.0 & 60.2 & \textbf{66.8} & \textbf{40.8} \\
\rowcolor{gray!15}\textit{Runtime} & & \textit{1447} & \textit{356} & \textit{1095} & \textit{777} & \textit{3675} & {695} \\
\midrule
\texttt{.jpg} & 4   & 49.7 & 69.4 & 64.0 & 51.5 & 58.7 & {27.8} \\
 & 8   & 55.7 & 71.4 & 64.8 & 55.1 & 61.8 & {31.3} \\
 & 16  & 58.9 & 70.6 & 63.6 & 56.9 & 62.5 & {37.0} \\
 & 32  & 61.2 & 70.1 & 62.2 & 57.2 & 62.7 & {37.1} \\
 & 64  & 62.6 & 68.3 & 61.5 & 58.1 & 62.6 & \textbf{38.9} \\
 & 128 & 62.7 & 66.8 & 60.6 & 57.5 & 61.9 & {36.5} \\
\rowcolor{gray!15}\textit{Runtime} & & \textit{2495} & \textit{663} & \textit{2329} & \textit{1337} & \textit{6824} & {1021} \\
\bottomrule
\end{tabular}}
\end{minipage}

\vspace{0.5em}
\begin{minipage}[t]{0.49\textwidth} 
\centering
\vspace{-5mm}
{\footnotesize(c)\cfgsampling}\\
\parbox{\linewidth}{\centering\tiny\textit{Fixed:} 602k px, HF, batch 1, torchcodec, \texttt{.mp4}, $T{=}0$}\\[3pt]
\resizebox{\SubWidth}{!}{%
\begin{tabular}{ll cccc | c | c}
\toprule
\SubHeadII{samp}{nf}
\midrule
Uniform & 32 & 61.4 & 73.3 & 68.6 & 58.7 & 65.5 & {36.1} \\
 & 64 & 63.5 & 74.2 & 69.0 & 60.2 & 66.7 & \textbf{39.2} \\
\rowcolor{gray!15}\textit{Runtime} & & \textit{542} & \textit{134} & \textit{413} & \textit{286} & \textit{1375} & {356} \\
\midrule
1\,fps & $\leq32$ & 61.5 & 69.9 & 64.1 & 57.7 & 63.3 & {34.9} \\
 & $\leq64$ & 63.5 & 69.9 & 64.0 & 58.7 & 64.0 & \textbf{39.4} \\
\rowcolor{gray!15}\textit{Runtime} & & \textit{542} & \textit{37.7} & \textit{135} & \textit{238} & \textit{953} & {241} \\
\bottomrule
\end{tabular}}
\end{minipage}
\hfill
\begin{minipage}[t]{0.49\textwidth}
\centering
{\footnotesize(d)\cfgdecoder}\\
\parbox{\linewidth}{\centering\tiny\textit{Fixed:} 602k px, uniform sampling, HF, batch 1, \texttt{.mp4}, $T{=}0$, 4 frames}\\[3pt]
\resizebox{\SubWidth}{!}{%
\begin{tabular}{l cccc | c | c }
\toprule
\SubHeadI{rdr}
\midrule
decord      & 50.1 & 64.5 & 59.4 & 50.4 & 56.1 & \textbf{25.2} \\
torchvision & 50.4 & 57.5 & 49.1 & 48.7 & 51.4 & {24.9} \\
torchcodec  & 50.1 & 64.4 & 59.2 & 50.3 & 56.0 & {24.8} \\
\rowcolor{gray!15}
\textit{Runtime} & \textit{100} & \textit{30.0} & \textit{104} & \textit{44} & \textit{278} & {97.9} \\
\bottomrule
\end{tabular}}
\end{minipage}

\vspace{0.5em}
\begin{minipage}[t]{0.49\textwidth}
\centering
{\footnotesize(e)\cfgresolution}\\
\parbox{\linewidth}{\centering\tiny\textit{Fixed:} uniform sampling, HF, batch 1, torchcodec, \texttt{.mp4}, $T{=}0$, 4 frames}\\[3pt]
\resizebox{\SubWidth}{!}{%
\begin{tabular}{l cccc | c | c}
\toprule
\SubHeadI{px}
\midrule
$200{,}704$ & 49.4 & 65.5 & 59.3 & 50.1 & 56.1 & {24.9} \\
$301{,}056$ & 50.1 & 64.5 & 59.0 & 50.5 & 56.0 & \textbf{25.2} \\
$602{,}112$ & 50.1 & 64.5 & 59.2 & 50.3 & 56.0 & {24.9} \\
\rowcolor{gray!15}\textit{Runtime} & \textit{82.5} & \textit{30.8} & \textit{101} & \textit{40.7} & \textit{255} & {95.3} \\
\bottomrule
\end{tabular}}
\end{minipage}
\hfill
\begin{minipage}[t]{0.49\textwidth}
\centering
\vspace{-4mm}
{\footnotesize(f)\cfgtemp}\\
\parbox{\linewidth}{\centering\tiny\textit{Fixed:} 602k px, uniform sampling, HF, batch 1, torchcodec, \texttt{.mp4}, 4 frames}\\[3pt]
\resizebox{\SubWidth}{!}{%
\begin{tabular}{l cccc | c | c}
\toprule
\SubHeadI{T}
\midrule
$0$   & 50.0 & 64.4 & 59.2 & 50.3  & 56.0 & {24.8} \\
$0.7$ & 47.1{\tiny$\pm$0.4} & 65.0{\tiny$\pm$3.0} & 56.9{\tiny$\pm$0.5} & 47.2{\tiny$\pm$1.1} & 54.0 & \textbf{25.9{\tiny$\pm$1.1}} \\
$0.8$ & 46.3{\tiny$\pm$1.3} & 63.9{\tiny$\pm$3.9} & 56.1{\tiny$\pm$0.8} & 45.9{\tiny$\pm$1.2} & 53.0 & {25.5{\tiny$\pm$1.4}} \\
$0.9$ & 45.6{\tiny$\pm$1.1} & 62.8{\tiny$\pm$3.5} & 55.6{\tiny$\pm$0.2} & 43.4{\tiny$\pm$0.0} & 51.9 & {25.3{\tiny$\pm$0.3}} \\
$1.0$ & 44.9{\tiny$\pm$0.3} & 61.5{\tiny$\pm$3.9} & 54.4{\tiny$\pm$0.2} & 43.3{\tiny$\pm$1.0} & 51.0 & {24.9{\tiny$\pm$1.2}} \\
\rowcolor{gray!15}\textit{Runtime} & \textit{172} & \textit{62.1} & \textit{169} & \textit{ 58.7} & \textit{461} & {169} \\
\bottomrule
\end{tabular}}
\end{minipage}

\label{tab:table2_allparameters}
\end{table*}
\subsection{Frame extraction}
\label{frame_extraction}

\cfgnframes and \cfgsampling[.] The number of input frames determines how much of a video the model sees and has one of the largest effects on accuracy. Uniform sampling selects exactly $N$ frames from a video of duration $T$, independent of $T$, so temporal coverage becomes sparser as duration $T$ increases. Fixed-rate sampling at $r$ frames per second selects approximately $rT$ frames, maintaining temporal density while allocating more frames to longer videos.

We find that the evaluated lmms\_eval implementation exposes a maximum frame count, $N_{\max}$, but its backends do not enforce it, which can cause out-of-memory errors on long videos. Our implementation enforces this limit: when it is set, the model receives $\min(\lfloor rT\rfloor, N_{\max})$ frames. If $\lfloor rT\rfloor \geq N_{\max}$, the framework switches to uniform sampling over the full video. Thus, fixed-rate sampling maintains temporal density until the cap is reached. We also support clipping the video based on $N_{\max}$; Appendix~\ref{appendix} provides details.

We observe that with uniform sampling, increasing the frame count from 4 to 128 raises accuracy on Video-MME from $50.0\%$ to $65.5\%$ and on \bench from $24.80\%$ to $40.75\%$ (Table~\ref{tab:table2_allparameters}(b)). Gains on TempCompass and MVBench saturate or reverse beyond 64 frames, showing that additional frames do not always improve performance. These results show that frame count is one of the strongest determinants of performance across the tested range of 4 to 128 uniformly sampled frames.

Sampling strategy also affects accuracy (Table~\ref{tab:table2_allparameters}(c)). Sampling at 1 fps instead of uniformly has no effect on Video-MME in this comparison. On the shorter clips in TempCompass and MVBench, it reduces both the number of frames and runtime, but lowers accuracy by $3.4$--$5.0$ pp.
Uniform sampling performs better on average, suggesting that dense sampling can help even on short videos.

\cfgdecoder[.] Video decoders such as decord~\citep{decord2019}, torchvision~\citep{torchvision2016}, and torchcodec~\citep{pytorch2024torchcodec} can differ in timestamp calculation, variable-frame-rate handling, and keyframe resolution. They may therefore extract different frames under the same sampling settings. 

In our evaluation, decoder choice changes scores but does not affect the overall model ranking (Table~\ref{tab:table2_allparameters}(d)). Decord performs slightly better than torchcodec, while torchvision produces larger drops on some datasets. We use torchcodec, the default in the evaluated \texttt{lmms\_eval} implementation, and recommend reporting the decoder to support reproducibility.



\cfgresolution[.] Before tokenization, frames are resized so that their area does not exceed \texttt{max\_pixels}. This budget controls spatial details and, for models with dynamic-resolution encoders, the number of visual tokens per frame. Different default budgets can therefore give the same model substantially different inputs at the same frame count. 

We evaluate the impact of the frame resolution strategy (Table~\ref{tab:table2_allparameters}(e)). We observe among the budgets tested, $200{,}704$ pixels yields the highest average accuracy across benchmarks, while $301{,}056$ performs best on \bench (Table~\ref{tab:table2_allparameters}(e)). However, the variation is minor and frame resolution does not impact performance significantly.


\enlargethispage{\baselineskip}
\subsection{Model input}
\label{model_input}
\cfgmodality[.] Sampled frames can be supplied as video input or as a sequence of images, and many VLMs process these modalities through different branches. The video branch may apply spatiotemporal pooling, merging patches across frames into shared visual tokens. The pooling scheme depends on the architecture: Qwen2.5-VL pools pairs of frames, producing half as many visual tokens as its image branch. In the implementations we evaluated, HF uses the video branch, whereas vLLM converts video inputs into individual images and bypasses temporal pooling. Appendix~\ref{appendix} describes our fix.
Table~\ref{tab:table2_allparameters}(b) compares direct \texttt{.mp4} input with pre-extracted \texttt{.jpg} frames. Input modality changes accuracy by $4$--$6$ percentage points in some settings. This recovers the gap between vLLM and HF in Table~\ref{tab:table2_allparameters}(a). Image input can perform better at low frame counts, gaining $5.0$ points on TempCompass at four frames, but its gains flatten or reverse at higher frame counts. At 128 frames on MVBench, image input scores $60.6\%$, compared with $68.0\%$ for video input. The lower visual-token count from spatiotemporal pooling may help explain the advantage of video input at larger frame budgets.

\cfgprompt[.] Evaluation prompts specify both the task and the required answer format, affecting predictions as well as whether those predictions are parsed correctly. Prompt layouts also vary: some place text before or after the video, while others repeat text or visual content to change which earlier inputs each modality can attend to under causal attention, resulting in prompt patterns such as T-V-T or V-T-V. We evaluate these choices in Section~\ref{promptdesign}.


\cfgtemp[.] Sampling temperature affects accuracy even when the prompt requests only a single option letter. Nonzero temperatures introduce stochasticity and can increase variation across runs. We observe from Table~\ref{tab:table2_allparameters}(f) that $\tau=0$ gives the best average accuracy among temperatures we test (except TempCompass and Meta-V which fall within the standard deviation). They also affect some datasets more than others: on Video-MME, accuracy falls from $50.0\%$ at $\tau=0$ to $44.9\%$ at $\tau=1$.



\subsection{Best Practices: Evaluation Cards}
\label{protocol}
As seen in above sections, VLMs are highly sensitive to evaluation choices, but are rarely reported in full. To make evaluations stable and reproducible, we fix a reference configuration and vary one factor, keeping all other settings fixed. Inspired by dataset cards~\citep{pushkarna2022datacard}, we introduce \emph{evaluation card} that summarizes the complete configuration behind its numbers. For each parameter varied within the table, all the other settings are fixed (Template in Table~\ref{tab:eval-card-template}). We provide an evaluation card for every table in this paper in Appendix~\ref{appendix}.
%
%
%
\makeatletter
\@ifpackageloaded{xcolor}{}{%
  \PackageError{evaluation_card}{xcolor is not loaded}%
  {Add \string\usepackage{xcolor} to your preamble.}}
\@ifpackageloaded{graphicx}{}{%
  \PackageError{evaluation_card}{graphicx is not loaded}%
  {Add \string\usepackage{graphicx} to your preamble.}}
\makeatother

\providecommand{\eccardwidth}{\linewidth}

\definecolor{cardblue}{HTML}{1A73E8}   
\definecolor{cardrule}{HTML}{C9D7EE}   

\ifdefined\ecgroupw\else\newlength\ecgroupw\fi
\ifdefined\ecfieldw\else\newlength\ecfieldw\fi
\ifdefined\ecvalw\else\newlength\ecvalw\fi

\providecommand{\ecgroup}[1]{\textcolor{cardblue}{\bfseries\textsc{#1}}}      
\providecommand{\echint}[1]{\leavevmode{\color{black!50}\itshape #1}}         
\providecommand{\ecsel}[1]{\textbf{#1}}                                        
\providecommand{\ecsep}{\unskip\ \textup{\textbar}\ \ignorespaces}            
\providecommand{\ecsweep}[1]{$\{$#1$\}$}                                       
\providecommand{\ecwrap}[1]{\raggedright #1\let\\\tabularnewline}              

\providecommand{\ectitle}[1]{%
  \multicolumn{3}{@{}l@{}}{\large\bfseries\color{cardblue}#1}\\
  \noalign{\vskip 2pt{\color{cardblue}\hrule height 0.8pt}\vskip 4pt}}
\providecommand{\ecbottomrule}{%
  \noalign{\vskip 3pt{\color{cardblue}\hrule height 0.8pt}}}
\providecommand{\ecgroupsep}{%
  \noalign{\vskip 3pt%
    \moveright\dimexpr\ecgroupw+2\tabcolsep\relax
    \vbox{\color{cardrule}\hrule width \dimexpr\linewidth-\ecgroupw-2\tabcolsep\relax height 0.4pt}%
    \vskip 3pt}}

\providecommand{\ecsetwidths}[2]{%
  \settowidth\ecgroupw{\ecgroup{#1}}%
  \settowidth\ecfieldw{#2}%
  \setlength\ecvalw{\dimexpr\linewidth-\ecgroupw-\ecfieldw-4\tabcolsep\relax}}

\begin{table*}[t]
\centering
\small
\renewcommand{\arraystretch}{1.15}
\ecsetwidths{Comments}{Inference backend}
\resizebox{0.99\eccardwidth}{!}{%
\begin{tabular}{@{}p{\ecgroupw}p{\ecfieldw}p{\ecvalw}@{}}
\ectitle{Evaluation Card  (Table \textit{ref})}
\ecgroup{Scope}    & Models            & Model-A-7B, Model-B-13B \\
                   & Datasets          & Dataset-1, Dataset-2 \\
\ecgroupsep
\ecgroup{Runtime}  & Inference backend & backend-name \\
                   & Batch size        & \ecsweep{1, 8, 32} \\
\ecgroupsep
\ecgroup{Input}    & Input modality    & .mp4 \\
                   & No.\ of frames    & \ecsweep{8, 16, 32} \\
                   & Frame sampling    & \ecsel{uniform}\echint{\ecsep fps rate} \\
                   & Max frames        & \ecsel{none}\echint{\ecsep int} \\
                   & Cap mode          & \echint{none\ecsep uniform\ecsep}\ecsel{cutoff} \\
\ecgroupsep
\ecgroup{Decoding} & Video decoder     & decoder-name \\
                   & Frame resolution  &  \echint{(max pixel count)} \\
                   & Temperature       & 0.0 \\
                   & Prompt layout     & \echint{vt\ecsep}\ecsel{tvt}\echint{\ecsep vtv} \\
                   & Prompt(s)         & \ecwrap{\texttt{<video> Answer the question using a single word or phrase. \{question\}}} \\
\ecgroupsep
\ecgroup{Comments} & Notes             & \ecwrap{Free-text remarks, e.g.\ deviations from the default setup, known issues, or hardware used (8$\times$A100).} \\
\ecbottomrule
\end{tabular}%
}
\vspace{-2mm}
\caption{\textbf{Evaluation card.} Reporting the settings that affects a score makes results reproducible and comparable across papers. Options in \textbf{bold} are the ones used; a set $\{\cdot\}$ denotes a parameter sweep.}
\label{tab:eval-card-template}
\vspace{-2mm}
\end{table*}

\section{Evaluation}

\subsection{Standardized Comparison of Baseline Models}
\label{subsec:eval_models}

\newcommand{\na}{--}   

\begin{table*}[t]
\centering
\caption{Re-evaluated vs.\ reported scores. For each model, the upper rows give
our own evaluation under a single fixed protocol as a function of the frame
budget, with all other settings held fixed; the shaded row at the bottom of each
block reproduces the numbers as reported in the literature, placed next to our
largest frame budget for direct comparison. Avg.\ is the mean over the seven
benchmarks; \textcolor{blue}{\textbf{Meta-V}} is highlighted in blue and reported
separately for short, medium and long videos, together with their unweighted mean.
\textbf{Bold} marks the best frame budget per model in each column.
\na{} denotes an entry that is not reported or not evaluated. Evaluation card is provided in~\ref{tab:eval-card-table4}.
}

\setlength{\aboverulesep}{0pt}
\setlength{\belowrulesep}{0pt}
\setlength{\extrarowheight}{1.5pt}   
\newcolumntype{B}{>{\columncolor{blue!8}}c}
\newcolumntype{E}{>{\columncolor{blue!8}[\tabcolsep][0pt]}c}  
\resizebox{\textwidth}{!}{%
\begin{tabular}{@{}lcccccccc|c|BBB|E@{}}
\toprule
\multirow{2}{*}{Model} & \multirow{2}{*}{\#Frames} &
Video- & Long- & Temp- & MMVU & LV- & \multirow{2}{*}{MLVU} & VSI- & \multirow{2}{*}{Avg.} &
\multicolumn{4}{>{\columncolor{blue!8}[\tabcolsep][0pt]}c@{}}{\textbf{\textcolor{blue}{Meta-V}}} \\
\cmidrule(l){11-14}


 & & MME & VideoBench & Compass & (Val) & Bench & & Bench & & Short & Medium & Long & Avg. \\
\midrule
\multirow{7}{*}{InternVL3.5-4B}
   &   4 & 54.5 & 52.1 & 69.1 & 46.4 & 38.0 & 59.5 & 46.3 & 52.3 & 35.4 & 24.2 & 32.3 & 30.6  \\
   &   8 & 57.7 & 54.5 & 68.9 & 46.7 & 38.9 & 60.6 & 50.6 & 54.0 & 45.0 & 32.7 & 35.1 & 37.6  \\
   &  16 & 60.7 & 56.8 & 71.1 & 48.6 & 41.7 & 63.4 & 55.1 & 56.8 & 50.4 & 36.2 & 37.9 & 41.5  \\
   &  32 & 63.1 & 58.3 & \textbf{71.3} & \textbf{50.8} & 42.6 & 67.1 & 56.8 & 58.6 & \textbf{53.9} & 41.2 & 41.5 & \textbf{45.5}  \\
   &  64 & \textbf{64.8} & \textbf{59.1} & 69.8 & 49.9 & 43.1 & 67.7 & \textbf{57.3} & \textbf{58.8} & 52.7 & 41.1 & 40.1 & 44.6  \\
   & 128 & 64.1 & 58.8 & 68.9 & 49.3 & \textbf{44.7} & \textbf{68.4} & 54.7 & 58.4 & 48.1 & \textbf{41.4} & \textbf{42.2} & 43.9  \\
\cmidrule(lr){2-14}
\rowcolor{gray!15}
   & rep.~\cite{wang2025internvl3_5}
         & 65.4 & 60.8 & \na & \na & \na & 70.4 & \na & \na & \na & \na & \na & \na \\
\midrule
\multirow{7}{*}{Qwen3-VL-4B}
   &   4 & 49.0 & 51.5 & 67.8 & 48.5 & 32.5 & 52.7 & 39.1 & 48.7 & 33.0 & 21.4 & 31.4 & 28.6  \\
   &   8 & 52.9 & 54.4 & 69.4 & 51.8 & 35.4 & 55.0 & 46.2 & 52.2 & 42.5 & 30.5 & 36.1 & 36.4  \\
   &  16 & 57.0 & 59.0 & 70.3 & 51.4 & 37.2 & 58.1 & 50.8 & 54.8 & 52.2 & 42.7 & 37.9 & 44.3  \\
   &  32 & 61.4 & 60.2 & \textbf{70.9} & 52.7 & 40.9 & 62.5 & 54.6 & 57.6 & 58.6 & 49.6 & 41.8 & 50.0  \\
   &  64 & 64.1 & 61.3 & 70.6 & 53.2 & 40.5 & 66.1 & 57.6 & 59.1 & 61.7 & 53.5 & \textbf{43.2} & 52.8  \\
   & 128 & \textbf{67.5} & \textbf{64.0} & 69.9 & \textbf{53.4} & \textbf{43.8} & \textbf{70.4} & \textbf{58.9} & \textbf{61.1} & \textbf{67.3} & \textbf{57.1} & 41.4 & \textbf{55.3}  \\
\cmidrule(lr){2-14}
\rowcolor{gray!15}
   & rep.~\cite{bai2025qwen3} & 69.3 & \na & \na & 50.5 & 56.2 & 75.3 & \na & \na & \na & \na & \na & \na \\ 
\midrule
\midrule
\multirow{7}{*}{Qwen2.5-VL-7B}
   &   4 & 50.1 & 50.3 & 64.5 & 46.2 & 28.1 & 51.9 & 26.8 & 45.4 & 26.7 & 22.7 & 25.1 & 24.8  \\
   &   8 & 53.4 & 52.7 & 70.4 & 48.4 & 33.1 & 54.5 & 30.5 & 49.0 & 30.2 & 26.6 & 27.9 & 28.2  \\
   &  16 & 58.1 & 56.2 & 71.8 & 50.8 & 34.7 & 56.3 & 33.5 & 51.6 & 35.1 & 36.0 & 29.0 & 33.4  \\
   &  32 & 61.4 & 58.7 & 73.3 & 49.7 & 37.4 & 59.5 & 36.7 & 53.8 & 42.1 & 33.3 & 29.3 & 34.9  \\
   &  64 & 63.5 & \textbf{60.2} & \textbf{74.2} & 51.8 & 39.3 & 63.4 & 37.7 & 55.7 & 44.9 & 40.1 & 33.6 & 39.5  \\
   & 128 & \textbf{65.6} & \textbf{60.2} & 73.4 & \textbf{51.9} & \textbf{42.1} & \textbf{66.1} & \textbf{38.3} & \textbf{56.8} & \textbf{46.1} & \textbf{41.3} & \textbf{34.3} & \textbf{40.6}  \\
\cmidrule(lr){2-14}
\rowcolor{gray!15}
   & rep.~\cite{bai2025qwen25vl}
         & 65.1 & 56.0 & 71.7 & 50.1 & 45.3 & 70.2 & \na & \na & \na & \na & \na & \na \\
\midrule
\multirow{7}{*}{InternVL3.5-8B}
   &   4 & 56.7 & 55.0 & 69.4 & 55.7 & 38.7 & 58.2 & 47.2 & 54.4 & 38.3 & 31.4 & 35.3 & 35.0  \\
   &   8 & 60.0 & 56.4 & 70.2 & 55.4 & 39.6 & 62.5 & 50.3 & 56.3 & 46.5 & 40.7 & 41.3 & 42.8  \\
   &  16 & 63.0 & 57.5 & 70.6 & 57.7 & 40.6 & 63.7 & 55.1 & 58.3 & 50.4 & 46.1 & 43.5 & 46.7  \\
   &  32 & 64.7 & 59.9 & \textbf{70.8} & 56.4 & 41.8 & 67.7 & \textbf{55.9} & 59.6 & 56.0 & 48.3 & 47.0 & 50.4  \\
   &  64 & \textbf{66.2} & \textbf{62.9} & 70.5 & \textbf{59.9} & 43.8 & \textbf{68.5} & 55.8 & \textbf{61.1} & \textbf{59.1} & 50.9 & 46.3 & 52.1  \\
   & 128 & 65.1 & 61.6 & 70.1 & 58.3 & \textbf{46.2} & \textbf{68.5} & 54.2 & 60.6 & 55.1 & \textbf{51.2} & \textbf{50.5} & \textbf{52.3}  \\
\cmidrule(lr){2-14}
\rowcolor{gray!15}
   & rep.~\cite{wang2025internvl3_5}
         & 66.0 & 62.1 & \na & \na & \na & 70.2 & \na & \na & \na & \na & \na & \na \\
\midrule
\multirow{7}{*}{Qwen3-VL-8B}
   &   4 & 53.0 & 52.2 & 70.7 & 53.8 & 33.6 & 53.0 & 41.4 & 51.1 & 37.1 & 28.8 & 41.7 & 35.9  \\
   &   8 & 56.1 & 55.9 & 72.3 & 56.7 & 36.0 & 54.7 & 47.5 & 54.2 & 44.6 & 36.9 & 44.2 & 41.9  \\
   &  16 & 60.3 & 58.0 & 73.2 & 58.4 & 37.4 & 59.2 & 51.4 & 56.8 & 56.4 & 44.7 & 49.8 & 50.3  \\
   &  32 & 64.6 & 59.5 & \textbf{74.4} & \textbf{59.5} & 39.8 & 63.1 & 55.8 & 59.5 & 61.5 & 51.9 & 51.6 & 55.0  \\
   &  64 & 66.5 & 63.0 & 73.2 & 58.5 & 42.9 & 68.2 & 59.0 & 61.6 & \textbf{65.5} & 55.8 & 50.9 & 57.4  \\
   & 128 & \textbf{69.2} & \textbf{63.6} & 74.2 & 58.7 & \textbf{46.5} & \textbf{72.1} & \textbf{59.7} & \textbf{63.4} & 62.5 & \textbf{60.9} & \textbf{53.4} & \textbf{58.9}  \\
\cmidrule(lr){2-14}
\rowcolor{gray!15}
   & rep.~\cite{bai2025qwen3}
         & 71.4 & \na & \na & 58.7 & 58.0 & 78.1 & \na & \na & \na & \na & \na & \na \\
\bottomrule
\end{tabular}%
}

\label{tab:table4_reported_vs_ours}
\end{table*}

Our ablations on the effect of different parameters show that evaluation settings can substantially change model scores. We therefore compared the performance of five common open MLLMs while controlling backend, resolution, temperature, and other settings. To assess the best model performance, we vary the frame budget, one of the strongest determinants of performance. This comparison reveals both how well models perform and how they use additional visual input.

\textbf{Setup.} We compare models at 4, 8, 16, 32, 64, and 128 frames on seven source benchmarks---Video-MME, LongVideoBench, TempCompass, MMVU, LVBench, and MLVU and VSI-Bench---and on \bench. The models are Qwen2.5-VL-7B~\cite{bai2025qwen25vl}, Qwen3-VL-4B-Instruct and Qwen3-VL-8B-Instruct~\cite{bai2025qwen3}, and InternVL3.5-4B and InternVL3.5-8B~\cite{wang2025internvl3_5}. Each InternVL3.5 frame is encoded as one tile. Table~\ref{tab:eval-card-table4} gives the full evaluation card; scores reported in the original model papers appear in gray.

\textbf{Results.} We report results in Table~\ref{tab:table4_reported_vs_ours}. Overall, it shows that Qwen3-VL achieves the best performance among the backbones in the 4B as well as in the 7-8B range for the maximum frame count (128 frames) setting. 

Comparing the resulting score with previously reported results, we see that on Video-MME, our results are similar to the originally reported scores. For Qwen2.5-VL-7B, our best configuration exceeds the reported scores on Video-MME, LongVideoBench, TempCompass, and MMVU. On LVBench, however, Qwen3-VL-4B scores $12.4$ points below its reported result. Sparse coverage of the benchmark's long videos under our 128-frame cap may explain this gap.

Frame count also changes model rankings. The Qwen models improve almost monotonically up to 128 frames, whereas InternVL3.5 peaks at 32--64 frames and then declines. 
Across the multiple source benchmarks, InternVL3.5-4B outperforms Qwen3-VL-4B at every frame count up to 32, including $52.3$ versus $48.7$ at four frames. The ranking reverses at 64 frames; at 128 frames, Qwen3-VL-4B leads by $2.7$ points ($61.1$ versus $58.4$). InternVL3.5-8B has the highest average score at 4--32 frames.

Meta-Video captures the broad performance trends observed across the source benchmarks: accuracy generally rises with frame count, while InternVL3.5-4B peaks earlier and then declines. These results support its use as a lower-cost comparison of the evaluated models and configurations.

\enlargethispage{\baselineskip}
\noindent\textbf{Frame count and video duration on \bench.} Table~\ref{tab:table4_reported_vs_ours} also reports performance on \bench by video duration. The Qwen models gain most between 4 and 32 frames. InternVL3.5-8B gains only $0.2$ points from 64 to 128 frames, while InternVL3.5-4B peaks at $45.5$ with 32 frames and falls to $43.9$ at 128. Short and medium videos benefit most from additional frames. Qwen3-VL-8B gains up to $32.1$ points on medium videos. Gains on long videos are smaller: Qwen2.5-VL-7B gains $9.2$ points on long videos, compared with $19.4$ on short videos. Even on short videos, however, accuracy saturates or declines beyond 32--64 frames for Qwen3-VL-8B and both InternVL3.5 models. Rankings on \bench also depend on the frame budget. InternVL3.5-4B outperforms Qwen3-VL-4B at four frames ($30.6$ versus $28.6$) and eight frames ($37.6$ versus $36.4$). From 16 frames onward, Qwen3-VL-4B leads, reaching $55.3$ versus $43.9$ at 128 frames. Qwen3-VL-4B also overtakes InternVL3.5-8B at 64 frames. Meaningful model comparisons therefore require controlled settings and an explicit frame budget.

\subsection{Evaluation of Prompt design}
\label{promptdesign}

\begin{table}[t]
\centering\small
\caption{\bench: Prompt and input modality ablation (.mp4 video vs.\ .jpg frames) with 4 frames. Average is the weighted mean of short, medium and long splits. Evaluation card is provided in Table~\ref{tab:eval-card-prompt-modality-ablation}.}
\resizebox{1.0\textwidth}{!}{%
\begin{tabular}{lccccccccc}
\toprule
\multirow{2}{*}{Prompt} & \multicolumn{2}{c}{Input modality} & \multirow{2}{*}{Layout} & \multirow{2}{*}{Number of frames} & \multirow{2}{*}{Max new token} & \multicolumn{4}{c}{\textcolor{blue}{\textbf{Meta-V}}} \\
\cmidrule(lr){2-3} \cmidrule(lr){7-10}
 & .mp4 & .jpg & & & & Small & Medium & Long & Avg. \\ \midrule
$P0$ (default)             & \cmark & \xmark & --  & 4 & default & 26.7 & 22.7 & 25.1 & 24.9\down{2.1} \\
$P1$ (think step-by-step)  & \cmark & \xmark & --  & 4 & 1024    & 17.8 & 27.7 & 30.7 & 24.8\down{2.26} \\
$P^*$ (single letter only) & \cmark & \xmark & --  & 4 & default & 26.0 & 27.4 & 27.9 & \textbf{27.0} \\
\midrule
$P0$ (default)             & \xmark & \cmark & VT  & 4 & default & 30.7 & 26.5 & 27.9 & 28.5\down{1.23} \\
$P1$ (think step-by-step)  & \xmark & \cmark & VT  & 4 & 1024    & 21.4 & 31.6 & 26.8 & 26.4\down{3.36} \\
$P^*$ (single letter only) & \xmark & \cmark & VT  & 4 & default & 30.7 & 30.4 & 27.5 & \textbf{29.7} \\
\midrule
$P0$ (default)             & \xmark & \cmark & TVT & 4 & default & 30.8 & 27.1 & 26.8 & 28.5\down{1.00} \\
$P1$ (think step-by-step)  & \xmark & \cmark & TVT & 4 & 1024    & 21.3 & 29.5 & 27.9 & 25.9\down{3.55} \\
$P^*$ (single letter only) & \xmark & \cmark & TVT & 4 & default & 29.5 & 31.9 & 26.5 & \textbf{29.5} \\
\midrule
$P0$ (default)             & \xmark & \cmark & VTV & 4 & default & 31.3 & 27.7 & 27.2 & 28.9\down{1.14} \\
$P1$ (think step-by-step)  & \xmark & \cmark & VTV & 4 & 1024    & 21.9 & 30.7 & 27.2 & 26.4\down{3.70} \\
$P^*$ (single letter only) & \xmark & \cmark & VTV & 4 & default & 29.5 & 31.6 & 28.9 & \textbf{30.1} \\
\bottomrule
\end{tabular}
}
\label{tab:mvb-prompt-modality-ablation}
\end{table}

\begin{table}[t]
\centering\small
\caption{\bench: Prompt and layout ablation across 32, 64 and 128 frames. All runs use .jpg frames as input; V and T denote the position of the visual input and the text prompt, respectively. Average is the weighted mean of short, medium and long splits; arrows show the difference to $P^*$ within each block. Evaluation card is provided in Table~\ref{tab:eval-card-prompt-layout-ablation}.}
\resizebox{1.0\textwidth}{!}{%
\begin{tabular}{lcccccccc}
\toprule
\multirow{2}{*}{Prompt} & \multirow{2}{*}{Layout} & \multirow{2}{*}{Number of frames} & \multirow{2}{*}{Max new token} & \multicolumn{4}{c}{\textcolor{blue}{\textbf{Meta-V}}} \\
\cmidrule(lr){5-8}
 & & & & Small & Medium & Long & Avg. \\ \midrule
$P0$ (default)             & VT  & 32 & default & 39.8 & 36.4 & 34.2 & 37.1\down{1.3} \\
$P1$ (think step-by-step)  & VT  & 32 & 1024    & 28.9 & 34.9 & 27.5 & 30.5\down{7.9} \\
$P^*$ (single letter only) & VT  & 32 & default & 37.9 & 42.3 & 34.2 & \textbf{38.4} \\
\midrule
$P0$ (default)             & TVT & 32 & default & 41.8 & 37.9 & 30.7 & 37.4\down{0.7} \\
$P1$ (think step-by-step)  & TVT & 32 & 1024    & 28.2 & 32.8 & 31.0 & 30.6\down{7.5} \\
$P^*$ (single letter only) & TVT & 32 & default & 37.1 & 43.5 & 32.8 & \textbf{38.1} \\
\midrule
$P0$ (default)             & VTV & 32 & default & 38.6 & 38.5 & 32.8 & 36.9\down{0.4} \\
$P1$ (think step-by-step)  & VTV & 32 & 1024    & 29.6 & 35.1 & 32.1 & 32.1\down{5.2} \\
$P^*$ (single letter only) & VTV & 32 & default & 34.7 & 42.0 & 35.3 & \textbf{37.3} \\
\midrule \midrule
$P0$ (default)             & VT  & 64 & default & 39.8 & 39.4 & 37.1 & 38.9\down{1.4} \\
$P1$ (think step-by-step)  & VT  & 64 & 1024    & 27.9 & 35.3 & 34.6 & 32.3\down{8.0} \\
$P^*$ (single letter only) & VT  & 64 & default & 39.4 & 45.0 & 36.0 & \textbf{40.3} \\
\midrule
$P0$ (default)             & TVT & 64 & default & 42.1 & 38.8 & 35.3 & \textbf{39.1}\up{2.0} \\
$P1$ (think step-by-step)  & TVT & 64 & 1024    & 23.9 & 31.6 & 30.3 & 28.3\down{8.8} \\
$P^*$ (single letter only) & TVT & 64 & default & 35.9 & 40.0 & 35.3 & 37.1 \\
\midrule
$P0$ (default)             & VTV & 64 & default & 39.2 & 36.4 & 32.5 & 36.4\down{2.1} \\
$P1$ (think step-by-step)  & VTV & 64 & 1024    & 29.2 & 33.7 & 32.5 & 31.6\down{6.9} \\
$P^*$ (single letter only) & VTV & 64 & default & 36.8 & 42.0 & 36.7 & \textbf{38.5} \\
\midrule \midrule
$P0$ (default)             & VT  & 128 & default & 37.5 & 36.4 & 35.3 & 36.5\down{1.7} \\
$P1$ (think step-by-step)  & VT  & 128 & 1024    & 27.1 & 36.1 & 32.8 & 31.7\down{6.5} \\
$P^*$ (single letter only) & VT  & 128 & default & 37.4 & 41.1 & 35.6 & \textbf{38.2} \\
\midrule
$P0$ (default)             & TVT & 128 & default & 40.6 & 36.1 & 32.8 & 36.9\down{0.4} \\
$P1$ (think step-by-step)  & TVT & 128 & 1024    & 28.0 & 33.7 & 33.5 & 31.5\down{5.8} \\
$P^*$ (single letter only) & TVT & 128 & default & 36.9 & 40.0 & 34.6 & \textbf{37.3} \\
\midrule
$P0$ (default)             & VTV & 128 & default & 36.7 & 32.2 & 28.6 & 32.9\down{3.1} \\
$P1$ (think step-by-step)  & VTV & 128 & 1024    & 27.9 & 32.2 & 32.8 & 30.7\down{5.3} \\
$P^*$ (single letter only) & VTV & 128 & default & 35.4 & 38.8 & 33.5 & \textbf{36.0} \\
\bottomrule
\end{tabular}
}
\label{tab:mvb-prompt-layout-ablation}
\end{table}
Second, we compare the impact of  different prompt wordings and architectures given a fixed setting. To this end, we consider both, additional instructions as well as text resp. video input repetitions. 

\noindent\textbf{Prompt wording.} Table~\ref{tab:mvb-prompt-modality-ablation} compares three prompts at a fixed budget of four frames. $P0$ is the dataset-specific default prompt in lmms\_eval. $P1$ asks for step-by-step reasoning: ``\texttt{Think step by step about what happens in the video, then end your reply with Answer:}''; its output is capped at 1{,}024 tokens. $P^*$ replaces the default instruction with ``\texttt{Respond with a single letter and nothing else.}'' Table~\ref{tab:eval-card-prompt-modality-ablation} gives the full evaluation card.

 $P^*$ achieves the highest average score for both input modalities, improving over $P0$ by $2.1$ points with \texttt{.mp4} input and $1.23$ points with \texttt{.jpg} input. Step-by-step prompting does not improve average accuracy: $P1$ scores $2.2$ points below $P^*$ with video input and more than $3.36$ points below it with image input.The effect varies by video duration. With video input, $P1$ performs best on long videos ($30.74$) but worst on short videos ($17.80$). At this four-frame budget, image input outperforms video input for every prompt. Appendix~\ref{appendix} provides further discussion.

\noindent\textbf{Prompt layout.} We also compare three orders of visual (V) and textual (T) content: visual input followed by text (VT), text before and after the visual input (TVT), and visual input before and after the text (VTV) for lower frame count of four frames in Table~\ref{tab:mvb-prompt-modality-ablation} and for higher frame counts (32, 64, 128 frames) in Table~\ref{tab:mvb-prompt-layout-ablation}. Table~\ref{tab:eval-card-prompt-layout-ablation} and Table~\ref{tab:eval-card-prompt-modality-ablation}, gives the evaluation card. $P^*$ performs best under every layout (except for TVT with 64 frame count as seen from Table~\ref{tab:mvb-prompt-layout-ablation}, where $P0$ performs better by 2\%) and reaches its highest score, $30.10$, with VTV. Layout has a smaller effect than wording: the average score for $P^*$ varies by only $0.6$ points ($29.50$--$30.10$), and the score for $P0$ varies by less than $0.5$ points. $P1$ performs worst under every layout, and repeating text or visual content does not consistently improve it. From Table~\ref{tab:mvb-prompt-layout-ablation}, at higher frame counts, (VT) layout performs the best, i.e. adding more textual or visual tokens doesn't help the model to obtain better performance.

\begin{figure*}[t]
	\centering
	\includegraphics[width=0.99\textwidth]{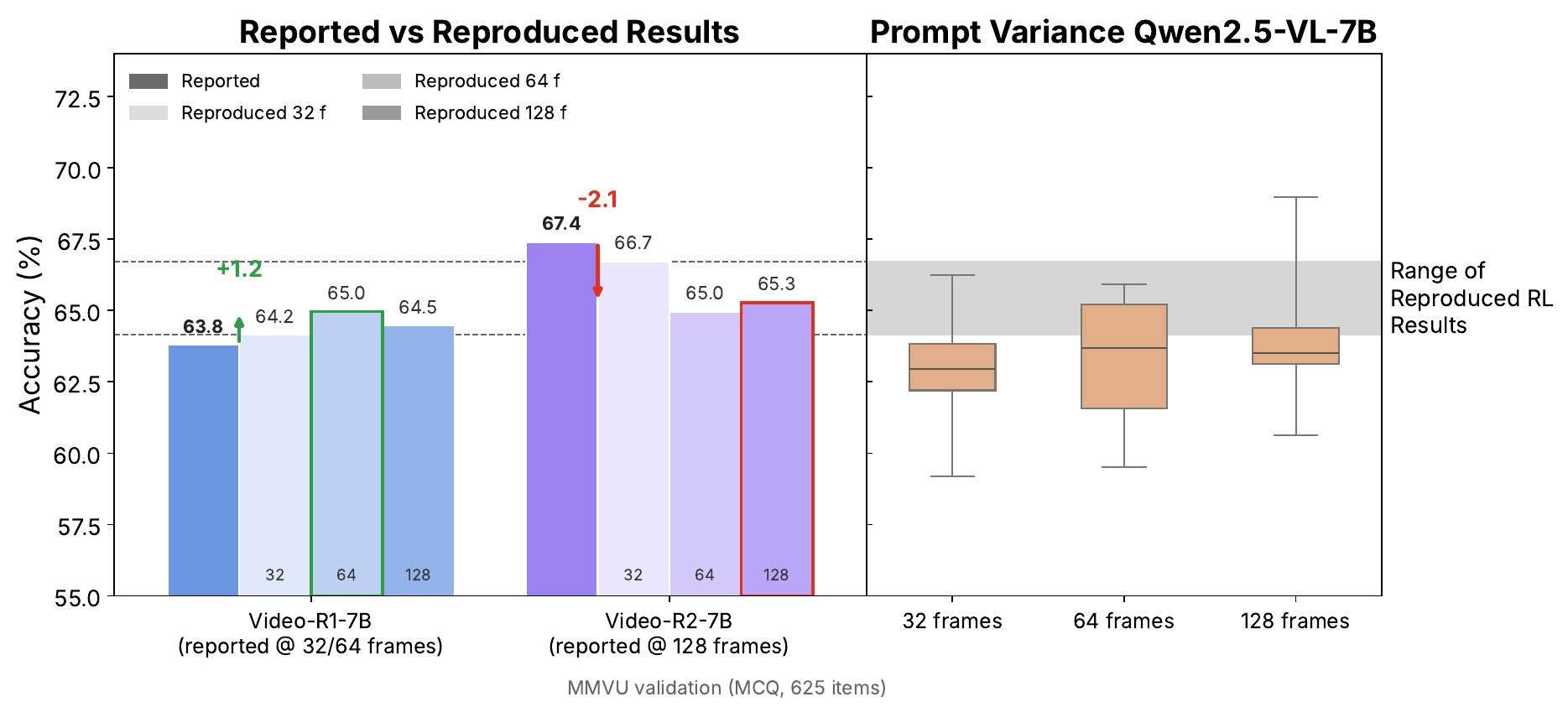}
	\caption{\textbf{Reported vs. reproduced MMVU accuracy.} Left: paper-reported scores (dark) vs. our reproductions at 32/64/128 frames (light). Right: prompt sensitivity of Qwen2.5-VL-7B across six prompts. The shaded band is the range of reproduced RL results, which falls within the base model’s prompt variance.
    }
	\label{fig:mmvu_reported_vs_measured}
\end{figure*}
\noindent\textbf{Comparison with Video reasoning models.} We re-evaluated two recent RL-based video reasoning models on the MMVU validation split (multiple-choice, 625 items). Each model was run with its own recommended prompt at 32, 64, 128 frames as shown in~\ref{fig:mmvu_reported_vs_measured} left. Out of the two models, only Video-R1 at 64 frames yields similar results compared to the reported numbers (64.96 reproduced vs reported 63.8). Video-R2 has 2.1 points less than reported (65.28 vs 67.4).

\vspace{-2mm}
\section{Conclusion}
\label{conclusion}
\vspace{-2mm}
We propose \bench, a compact video understanding benchmark that is an efficient and reliable version of the source benchmarks. This benchmark helps us to evaluate models efficiently while still keeping the trends of the original benchmarks. We propose an open evaluation framework built on top of \texttt{lmms\_eval} for cleaner video understanding
evaluation. We provide an exhaustive evaluation with various configurations on seven benchmarks. 
We provide an analysis of the effect of input layout and varying prompts on \bench.
In conclusion, we show that evaluation choices such as frame sampling, input layout, prompt, backend affects video benchmark scores, and are often unreported. We hope the new benchmark and evaluation reporting will help to pave the way towards more transparent, reproducible evaluation in the field.

\subsubsection*{Acknowledgments}
Soumya Jahagirdar is funded by the European Research Council (ERC) under the Starting Grant GraViLa (101117556). This work is supported by the Tübingen AI Center. This research is in part supported by the MIT-IBM Watson AI Lab. Ameya Prabhu acknowledges funding by the Federal Ministry of Research, Technology and Space (BMFTR), FKZ: 16IS24085B. Vishaal Udandarao was supported by a Google PhD Fellowship in Machine Intelligence. The authors gratefully acknowledge the Gauss Centre for Supercomputing e.V. \nolinkurl{www.gauss-centre.eu} Alvarez (2021) for funding this work by providing computing time on the GCS Supercomputer JUPITER—JUWELS at the Jülich Supercomputing Centre (JSC). This work also acknowledges support from the project ELLIOT-FM: Open Multi-Modal Foundation Models with Strong Generalization and Reasoning.

\bibliography{iclr2027_conference}
\bibliographystyle{plainnat}
\clearpage
\appendix
\part*{Appendix}
\addcontentsline{toc}{part}{Appendix}  
\etocsetnexttocdepth{2}
\localtableofcontents

\section{Appendix}
\label{appendix}

\subsection{\texorpdfstring{\cfgmodality}{Modality}}

\textbf{How} \texttt{.mp4} \textbf{is processed?}
For inputs with .mp4, the processor first samples frames based on \texttt{nframes} argument. It then resizes the frames based on the max\_pixel. Further, frames are grouped in pairs due to temporal patch size which is specific for videos in Qwen2.5-VL models. Each frame is encoded separately to obtain $X_{vision}$ tokens per frame, meaning that there are $\texttt{nframes} \times X_{vision}$ tokens. Then, for each 2x2 group of adjacent patch embeddings (based on the temporal patch size), a spatial merge happens. The processor inserts \texttt{<|vision\_start|>}, then one \texttt{<|video\_pad|>} with token id $151656$ per merged token, then \texttt{<|vision\_end|>}. So the total video tokens is $(\frac{\texttt{nframes}}{2}) \times \frac{H}{28} \times \frac{W}{28}$. It also uses M-RoPE with separate temporal, height, and width components.

\textbf{How} \texttt{.jpg} \textbf{is processed?}
The wrapper converts the PIL image to data and resize based on \texttt{max\_pixel} to multiples of 28. Instead of the true temporal dimension which was added in the case of .mp4 as input, in .jpg, each frame is duplicated, followed by similar path as video processing but since the tokens are duplicated, the total token count become $\frac{H}{28} \times \frac{W}{28}$ with M-RoPE temporal index as constant for all image tokens. The image placeholder id is $151655$.

\subsection{\texorpdfstring{\cfgbackend}{Modality}}

Similar to \texttt{.mp4} and \texttt{.jpg}, the choice of inference backend \texttt{HF} vs \texttt{vLLM} also affects results because the two differ in how they process visual inputs. \texttt{HF} backend accepts video natively along with images, whereas \texttt{vLLM} by default, samples frames and encodes the frames as separate images. As a result, there is no spatio-temporal pooling with \texttt{vLLM}, which leads to difference in performance based on the backend used.

\subsection{Original vLLM vs fixed vLLM}
The issue with the original vLLM backend is that it decodes video with qwen\_vl\_utils and sends the sampled frames to vLLM as N separate images, so Qwen2.5-VL receives N \texttt{<|image\_pad|>} blocks with no two-frame temporal merge, no video position ids, i.e., the video silently falls back into frames condition. 
The fixed backend, passes the .mp4 as a file (\texttt{pass\_video\_url=True} in the function builder), so vLLM decodes it through the model's video processor and produces \texttt{<|video\_pad|>} tokens, matching the input used in HF backend. In Table~\ref{tab:vllm_fix}, we show the comparison of the HF runs, vLLM original scores (boosted scores), and vLLM fixed scores. It can be seen that the scores using the vLLM fixed version are closer to HF.

\begin{table}[t]
\centering
\caption{In this table, we provide the scores of \metav after re-evaluating with the fixed vLLM, and compare it with the existing vLLM scores.}
\begin{tabular}{@{}c|cccc@{}}
\toprule
Batch size & HF & vLLM original & vLLM fixed & $\Delta$ \\
\midrule
1 & 24.81 & 31.00 & 26.25 & $-4.75$ \\
4 & 24.34 & 30.52 & 26.16 & $-4.35$ \\
8  & 24.01 & 30.62 & 26.86 & $-3.77$ \\
16 & 24.61 & 30.40 & 26.66 & $-3.74$ \\
32 & 24.62 & 30.47 & 27.22 & $-3.26$ \\
\bottomrule
\end{tabular}
\label{tab:vllm_fix}
\end{table}

\subsection{Uniform vs 1fps}

As explained in the main paper, we re-implement 1fps as the original lmms\_eval lacked the full implementation. We make a design choice when using 1fps. In the original lmms\_eval, if the $N_{max}$ argument was provided, it would ignore it for the cases when the duration of the video was longer than $N_{max}$, and would result in out of memory error. Hence, we re-implement this. In our implementation, if the 1fps is used along with $N_{max}$ argument, all the videos which are less than $N_{max}$ will be sampled at 1fps, but for the videos longer than $N_{max}$, a fallback of uniform $N_{max}$ is done. One other way to implement this is by clipping all the frames that come after $N_{max}$. By doing so, we might end up not seeing the full video, and it could happen that the question asked is from the part of the video which was clipped off. Due to this issue we decided on using uniform $N_{max}$ sampling as the fallback. To support this claim, on randomly sampled 100 videos from VideoMME long subset, we perform the clipped version vs uniform sampled version. The uniform sampled version with $N_{max}$=32 obtains $49.33$ score whereas the clipped version obtains $47.33$ score which is $2.0$ points lower. Hence we decided on uniform sampled version as the fallback. With 64 frames the uniform version is $52.33$ score, whereas with the clipped version, we get $50.0$ as score.

\subsection{Additional details about \metav}

\begin{figure*}[t]
	\centering
	\includegraphics[width=.75\textwidth]{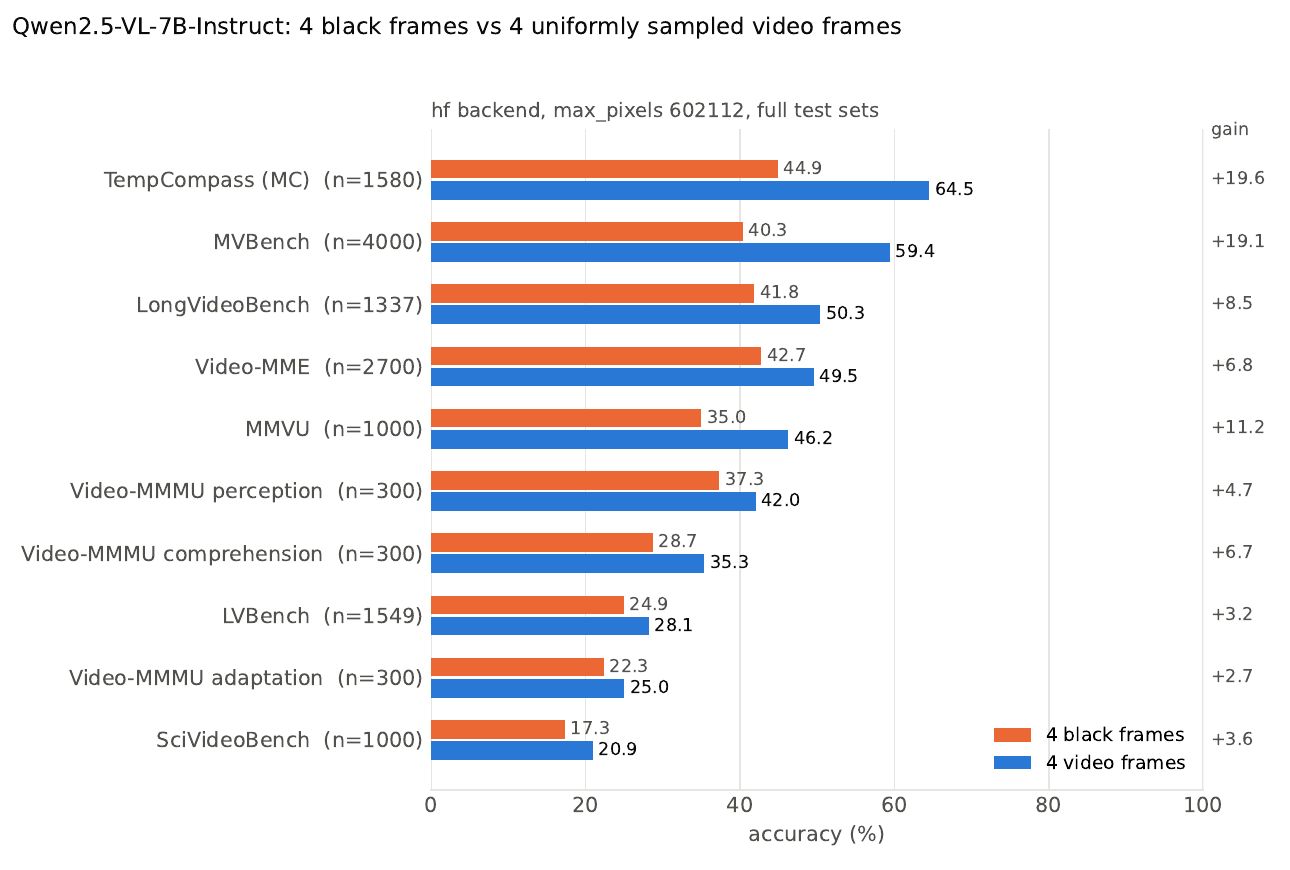}
	\caption{Blind baseline performance vs. four uniformly sampled frames, for Qwen2.5-VL-7b
    }
	\label{fig:black_vs_video_qwen25vl7b}
\end{figure*}

\newcommand{\nq}[1]{{\scriptsize\textcolor{gray}{#1}}}

\begin{table*}[t]
\centering
\caption{\textbf{Blind baseline.} Scores when every input frame is replaced by a
black frame, so the only usable signal is the question and answer text. All runs
use 4 frames, HuggingFace inference, batch size 1, temperature 0;
\texttt{max\_pixels}$=602{,}112$ for the Qwen models and
\texttt{max\_patches}$=12$ for InternVL3.5. \#Q is the number of questions
evaluated.}
\resizebox{\textwidth}{!}{%
\begin{tabular}{@{}lcccccccccc@{}}
\toprule
\multirow{2}{*}{Model} & Video- & Long- & Temp- & MMVU & LV- & MV- &
\multicolumn{3}{c}{VideoMMMU} & SciVideo- \\
\cmidrule(lr){8-9}
 & MME & VideoBench & Compass & (Val) & Bench & Bench &
Perc. & Comp. & Adapt. & Bench  \\
\nq{\#Q} & \nq{2700} & \nq{1337} & \nq{1580} & \nq{1000} & \nq{1549} & \nq{4000} &
\nq{300} & \nq{300} & \nq{300} & \nq{1000} \\
\midrule
Qwen2.5-VL-7B  & 42.7 & 41.8 & 44.9 & 35.0 & 24.9 & 40.3 & 37.3 & 28.7 & 22.3 & 17.3 \\
Qwen3-VL-8B    & 39.0 & 41.6 & 37.9 & 39.7 & 15.2 & 31.2 & 26.7 & 29.7 & 26.3 & 23.1 \\
InternVL3.5-8B & 43.6 & 45.9 & 41.7 & 39.4 & 26.7 & 35.3 & 35.0 & 31.0 & 30.3 & 23.9 \\
\bottomrule
\end{tabular}%
}
\vspace{2pt}
\begin{minipage}{\textwidth}
\footnotesize
\end{minipage}

\label{tab:blind_baseline}
\end{table*}

In Figure~\ref{fig:metav-benchmark-stats}, we show the statistics about our \metav dataset. \metav contains 1000 QA pairs on 788 unique videos (but 791 distinct videos in total), which are obtained from five source benchmarks. The source benchmarks are: VideoMME, LongVideoBench, LVBench, VideoMMMU, and VSIBench. In order to get a cleaner set of QA pairs, we first do a blind filtering. In this step, three models: Qwen2.5-VL-7b, Qwen3-VL-8b-Instruct and InternVL3.5-8b, were give with black frame as input instead of the video frames. In Table~\ref{tab:blind_baseline}, we show the results, where each row shows the accuracy that can be obtained without actually looking at visual tokens. In Figure~\ref{fig:black_vs_video_qwen25vl7b}, we also show the performance comparison for a blind model that sees black frames, and compare it with the same model with four uniformly sampled frames.  Ideally, such questions contribute to the noise and do not help in assessing the capabilities of video understanding models. Few videos in LVBench and VideoMME that were extremely long (i.e., $\ge$ 60 minutes) were also removed. From the remaining pool, the questions were sampled such that, primary category of every source is present in \metav which includes: Video-MME: 101 short / 109 medium / 102 long; all 12 task types, 6 domains, 30 sub-categories; LongVideoBench: all 17 question categories with 10 to 23 items each; duration groups 15 s / 60 s / 600 s / 3600 s with 38 / 49 / 72 / 81 items; 10 topic categories; LVBench: 6 question types, dominated by entity recognition (57) and event understanding (48), 6 genres; VideoMMMU: 60 per track, all 30 disciplines with 3 to 9 items, 6 QA types, all multiple choice; VSI-Bench: all 10 question types, 55 ScanNet / 42 ARKitScenes / 25 ScanNet++ scenes. The following is the spread of (\#Videos,\#QA pairs) in each split; Short - (382 QA pairs, 335 videos), Medium - (335 QA pairs, 281 videos), Long - (283 QA pairs, 172 videos)
Two types of scores can be obtained for \metav and each of them is specified in the tables: (a) weighted mean of short, medium, and long videos, where the score is weighted by the number of samples in each bucket, (b) non-weighted mean of short, medium, and long videos, which is the direct mean of the three scores.


\subsection{Base models vs Reasoning models}
\providecolor{dgreen}{RGB}{0,131,0}
\providecolor{dred}{RGB}{200,40,40}

\begin{table}[t]
\centering
\caption{VideoMME (2{,}700 questions) accuracy (\%) vs.\ number of uniformly sampled frames. All reasoning models are fine-tuned from Qwen2.5-VL-7B-Instruct; parentheses give the difference to the base model. 
}
\small
\setlength{\tabcolsep}{4pt}
\resizebox{\linewidth}{!}{%
\begin{tabular}{lcccccc}
\toprule
 & \multicolumn{6}{c}{\#Frames} \\
\cmidrule(lr){2-7}
Model & 4 & 8 & 16 & 32 & 64 & 128 \\
\midrule
Qwen2.5-VL-7B & 50.1 & 53.4 & 58.1 & \textbf{61.4} & \textbf{63.5} & \textbf{65.6} \\
\midrule
Video-R1-7B & 48.5 \,{\scriptsize\textcolor{dred}{(-1.6)}} & 52.0 \,{\scriptsize\textcolor{dred}{(-1.4)}} & 56.3 \,{\scriptsize\textcolor{dred}{(-1.8)}} & 60.0 \,{\scriptsize\textcolor{dred}{(-1.4)}} & 61.5 \,{\scriptsize\textcolor{dred}{(-2.0)}} & 62.5 \,{\scriptsize\textcolor{dred}{(-3.1)}} \\
VideoChat-R1.5-7B & \textbf{51.7} \,{\scriptsize\textcolor{dgreen}{(+1.6)}} & \textbf{54.1} \,{\scriptsize\textcolor{dgreen}{(+0.7)}} & \textbf{58.8} \,{\scriptsize\textcolor{dgreen}{(+0.7)}} & 61.4 \,{\scriptsize\textcolor{dgreen}{(0.0)}} & 63.4 \,{\scriptsize\textcolor{dred}{(-0.1)}} & 63.7 \,{\scriptsize\textcolor{dred}{(-1.9)}} \\
Video-R2-7B & 50.1 \,{\scriptsize\textcolor{dgreen}{(0.0)}} & 53.4 \,{\scriptsize\textcolor{dgreen}{(0.0)}} & 57.6 \,{\scriptsize\textcolor{dred}{(-0.5)}} & 61.3 \,{\scriptsize\textcolor{dred}{(-0.1)}} & 63.4 \,{\scriptsize\textcolor{dred}{(-0.1)}} & {64.9} \,{\scriptsize\textcolor{dred}{(-0.7)}}\\
\bottomrule
\end{tabular}}

\label{tab:videomme_base_vs_reasoning_frames}
\end{table}

\begin{table}[t]
\centering
\caption{Accuracy (\%) on MMVU (validation split, MCQ, 625 items) and SciVideoBench (test split, 1000 items) with 32, 64, and 128 input frames. Best result per column in bold.}
\label{tab:mmvu_scivideobench}
\begin{tabular}{llcccccc}
\toprule
 & & \multicolumn{3}{c}{MMVU} & \multicolumn{3}{c}{SciVideoBench} \\
\cmidrule(lr){3-5} \cmidrule(lr){6-8}
Model & Prompt & 32 & 64 & 128 & 32 & 64 & 128 \\
\midrule
\multirow{6}{*}{Qwen2.5-VL-7B}
 & default              & 63.36 & 65.92 & 64.64 & 26.1 & 26.9 & 27.9 \\
 & CoT         & 66.24 & 65.28 & \textbf{68.96} & 24.4 & 23.6 & 25.6 \\
 & video r1             & 64.00 & 64.96 & 63.04 & 23.4 & 26.1 & 26.7 \\
 & video r2             & 62.08 & 61.28 & 63.36 & 23.6 & 25.3 & 25.5 \\
 & video chat r15 think & 59.20 & 59.52 & 60.64 & 24.0 & 27.3 & 26.7 \\
 & video chat r15 qa    & 62.56 & 62.40 & 63.68 & 25.6 & 26.2 & 27.3 \\
\midrule
\multirow{2}{*}{Video-R1-7B}
 & default              & 55.20 & 53.76 & 55.52 & 22.4 & 25.0 & 26.4 \\
 & video r1             & 64.16 & 64.96 & 64.48 & 26.7 & \textbf{28.8} & 27.9 \\
\midrule
\multirow{2}{*}{Video-R2}
 & default              & 63.84 & 64.16 & 64.80 & 25.8 & 27.7 & \textbf{28.6} \\
 & video r2             & \textbf{66.72} & 64.96 & 65.28 & 27.8 & 26.3 & 27.7 \\
\midrule
\multirow{2}{*}{VideoChat-R1.5-7B}
 & default              & 64.48 & \textbf{66.24} & 64.16 & 26.8 & 27.8 & 28.4 \\
 & video chat r15 think & 64.80 & 64.96 & 65.12 & \textbf{28.3} & 27.8 & 27.4 \\
\bottomrule
\end{tabular}
\end{table}
Recently, a lot of effort in video understanding community has shifted towards building better video reasoning models. Video-R1 \cite{feng2025video} introduces a T-GRPO algorithm, specifically designed to handle temporal information in videos. It utilizes two datasets: Video-R1-CoT-165k for supervised finetuning (SFT) and Video-R1-260k for reinforcement learning (RL) training. Video-RFT \cite{wang2025videorft} proposes a multi-expert driven, cognition-inspired CoT curation pipeline. In this framework, an LLM first generates preliminary CoTs based on rich, structured, and literal representations of video content. A VLM then refines these CoTs by conditioning them on the actual video input. This process results in two datasets: VideoRFT-CoT-102K for SFT and VideoRFT-RL-310K for RL training. In contrast, Video-RTS \cite{wang2025videorts} presents a different approach by combining efficient RL with a video-adaptive test-time scaling (TTS) strategy. All of these works use Qwen2.5-VL-7b as the base model and continue training on this. 

In Table~\ref{tab:videomme_base_vs_reasoning_frames}, we show the performance of Video-R1, Video-R2, and VideoChat-R1.5 reasoning models on VideoMME~\cite{fu2025videomme} dataset. It can be seen that the reasoning models provide only marginal gains over their base model, raising the concerns of necessity of such models. In Table~\ref{tab:mmvu_scivideobench}, we show the performance of MMVU (validation split, MCQ) and SciVideoBench on these three models. It can be seen that on MMVU, the highest score across all models, prompts and frame counts is achieved by the base Qwen2.5-VL-7B with a CoT prompt (68.96 at 128 frames), exceeding the best reasoning-model result (66.72, Video-R2 at 32 frames) . For SciVideoBench, the reasoning model result (Video-R1 with its own prompt, 28.8 at 64 frames) exceeds the best base-model result (27.9) by only 0.9 points.

\begin{figure*}[t]
	\centering
	\includegraphics[width=.85\textwidth]{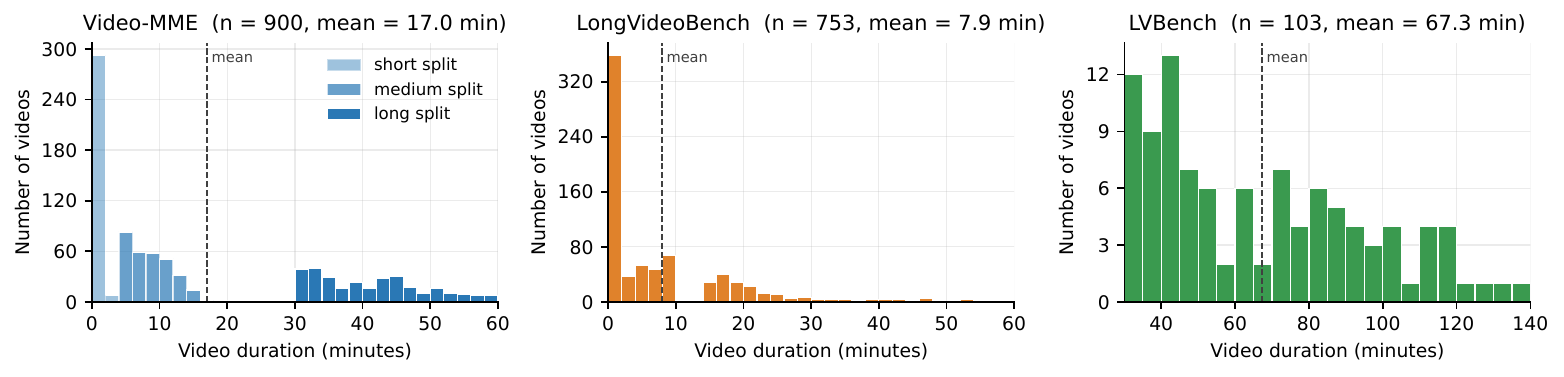}
	\caption{{Video duration of VideoMME~\citep{fu2025videomme}, LongVideoBench~\citep{wu2024longvideobench}, LVBench~\citep{wang2025lvbench}}. 
    }
	\label{fig:duration_longvideobench_lvb_videommelong}
\end{figure*}
\subsection{Video duration of long video benchmarks.}
As mention in section~\ref{subsec:metav_stats}, different benchmarks have different definitions of long videos. From Figure~\ref{fig:duration_longvideobench_lvb_videommelong}, it can be seen that the average duration of the long split of VideoMME dataset is 41 minutes (17.0 over all splits: short, medium and long), whereas for LongVideoBench it is 7.9 minutes, and for LVBench it is 67.3 minutes. For \metav, we divide the videos into three splits based on the duration of the video as explained in section~\ref{subsec:metav_stats}.

\subsection{Evaluation datasets and settings}
The following are the datasets are their splits used for evaluation.

\textbf{VideoMME.} We evaluate \texttt{videomme} without subtitles. This benchmark has 2700 QA pairs, and 900 videos.
The default prompt of this dataset is:
\begin{verbatim}
Select the best answer to the following multiple-choice question based on
the video. 
Respond with only the letter (A, B, C, or D)
of the correct option.
<question>
A. <option>
B. <option>
C. <option>
D. <option>

Answer with the option's letter from the given choices directly.
\end{verbatim}

\textbf{LongVideoBench.} We evaluate \texttt{longvideobench\_val\_v}, which contains 1337 QA pairs,and 753 videos.
The default prompt of this dataset is:
\begin{verbatim}
<question>
A. <option0>
B. <option1>
C. <option2>
D. <option3>
E. <option4>
Answer with the option's letter from the given choices directly.
\end{verbatim}

\textbf{TempCompass.} We evaluate \texttt{tempcompass\_multi\_choice}, which contains 1580 QA pairs, and 410 videos.
The default prompt of this dataset is:
\begin{verbatim}
<question>
A. <option>
B. <option>
C. <option>
Please directly give the best option:
\end{verbatim}

\textbf{MMVU.} We evaluate \texttt{mmvu\_val}, which contains 1000 QA pairs, and 583 videos.
The default prompt of this dataset is:
\begin{verbatim}
Question:<question>
A: <a>
B: <b>
C: <c>
D: <d>
E: <e>
Visual Information: processed video
Do not generate any intermediate reasoning process. 
Answer directly with the option letter from the given choices.
\end{verbatim}
For Open-ended:

\begin{verbatim}
Question:<question>
Visual Information: processed video
Do not generate any intermediate reasoning process. 
Directly output the final answer.
\end{verbatim}

\textbf{LVBench.} We evaluate \texttt{lvbench}, which contains 1549 QA pairs, and 103 videos.
The default prompt of this dataset is:
\begin{verbatim}
<question>
(A) <option>
(B) <option>
(C) <option>
(D) <option>
Answer the question with the option letter
\end{verbatim}

\textbf{VSIBench.} We evaluate \texttt{vsibench}, which contains 5130 QA pairs on 288 videos.
The default prompt of this dataset is:
\begin{verbatim}
These are frames of a video.
<question>
Options:
<option A line>
<option B line>
...
Answer with the option's letter from the given choices directly.
\end{verbatim}
For numerical question types:
\begin{verbatim}
These are frames of a video.
<question>
Please answer the question using a single word or phrase.
\end{verbatim}

\textbf{MLVU.} We evaluate \texttt{mlvu\_dev}, which contains 2174 QA pairs on 1222 videos.
The default prompt of this dataset is:
\begin{verbatim}
<question>
(A) <option>
(B) <option>
(C) <option>
(D) <option>

Only give the best option.
Best option: 
\end{verbatim}

\subsection{Evaluation cards}

We provide evaluation cards as a template which can be provided with every evaluation table for better reproducibility.
%
%
\makeatletter
\@ifpackageloaded{xcolor}{}{%
  \PackageError{evaluation_card}{xcolor is not loaded}%
  {Add \string\usepackage{xcolor} to your preamble.}}
\makeatother

\definecolor{cardblue}{HTML}{1A73E8}   
\definecolor{cardrule}{HTML}{C9D7EE}   

\ifdefined\ecgroupw\else\newlength\ecgroupw\fi
\ifdefined\ecfieldw\else\newlength\ecfieldw\fi
\ifdefined\ecvalw\else\newlength\ecvalw\fi

\providecommand{\ecgroup}[1]{\textcolor{cardblue}{\bfseries\textsc{#1}}}      
\providecommand{\echint}[1]{\leavevmode{\color{black!50}\itshape #1}}         
\providecommand{\ecsel}[1]{\textbf{#1}}                                        
\providecommand{\ecsep}{\unskip\ \textup{\textbar}\ \ignorespaces}            
\providecommand{\ecsweep}[1]{$\{$#1$\}$}                                       
\providecommand{\ecwrap}[1]{\raggedright #1\let\\\tabularnewline}                           

\providecommand{\ectitle}[1]{%
  \multicolumn{3}{@{}l@{}}{\large\bfseries\color{cardblue}#1}\\
  \noalign{\vskip 2pt{\color{cardblue}\hrule height 0.8pt}\vskip 4pt}}
\providecommand{\ecbottomrule}{%
  \noalign{\vskip 3pt{\color{cardblue}\hrule height 0.8pt}}}
\providecommand{\ecgroupsep}{%
  \noalign{\vskip 3pt%
    \moveright\dimexpr\ecgroupw+2\tabcolsep\relax
    \vbox{\color{cardrule}\hrule width \dimexpr\linewidth-\ecgroupw-2\tabcolsep\relax height 0.4pt}%
    \vskip 3pt}}

\providecommand{\ecsetwidths}[2]{%
  \settowidth\ecgroupw{\ecgroup{#1}}%
  \settowidth\ecfieldw{#2}%
  \setlength\ecvalw{\dimexpr\linewidth-\ecgroupw-\ecfieldw-4\tabcolsep\relax}}

\providecommand{\ectodo}[1][to fill in]{\textcolor{red!70!black}{\itshape[#1]}}  
\providecommand{\na}{--}
\providecommand{\ecprompt}[1]{{\footnotesize\ttfamily\frenchspacing #1}}      
\providecommand{\ecQ}{\echint{\textrm{$\langle$Q$\rangle$}}}                  
\providecommand{\ecO}{\echint{\textrm{$\langle$O$\rangle$}}}                  
\providecommand{\ecalt}[1]{\echint{\textrm{#1:}}}                             

\begin{table*}[t]
\centering
\small
\renewcommand{\arraystretch}{1.15}
\ecsetwidths{Comments}{Inference backend}
\begin{tabular}{@{}p{\ecgroupw}p{\ecfieldw}p{\ecvalw}@{}}
\ectitle{Evaluation Card \textmd{\normalsize\color{black!60}(Table~\ref{tab:table4_reported_vs_ours})}}
\ecgroup{Scope}    & Models            & \ecwrap{InternVL3.5-4B, Qwen3-VL-4B, Qwen2.5-VL-7B, InternVL3.5-8B, Qwen3-VL-8B} \\
                   & Datasets          & \ecwrap{Video-MME, LongVideoBench, TempCompass, MMVU (val), LVBench, VSI-Bench, MLVU, Meta-V (short / medium / long)} \\
\ecgroupsep
\ecgroup{Runtime}  & Inference backend & HF \\
                & Batch size        & \ecwrap{\echint{per no.\ of frames:} 4\,$\to$\,16, 8\,$\to$\,8, 16\,$\to$\,4, 32\,$\to$\,4, 64\,$\to$\,2, 128\,$\to$\,1} \\
\ecgroupsep
\ecgroup{Input}    & Input modality    & .mp4 \\
                   & No.\ of frames    & \ecsweep{4, 8, 16, 32, 64, 128} \\
                   & Frame sampling    & \ecsel{uniform} \\
                   & Max frames        & \ecsel{none}\\
                   & Cap mode          & \ecsel{none} \\
\ecgroupsep
\ecgroup{Decoding} & Video decoder     & torchcodec \\
                   & Frame resolution  & 602112 \echint{(max pixel count)} \\
                   & Temperature       & 0.0 \\
                   & Prompt layout     & \ecsel{vt} \\
                   & Prompt(s)         & \ecwrap{Dataset default prompts \echint{(see Comments)}} \\
\ecgroupsep
\ecgroup{Comments} & Protocol          & \ecwrap{One fixed protocol for all models; only the frame budget varies, all other settings are held fixed.} \\
                   & Reported rows     & \ecwrap{Shaded ``rep.'' rows are copied from the original papers, not re-run; \na{} = not reported or not evaluated.} \\
                   & Aggregates        & \ecwrap{Avg.\ = mean over the seven benchmarks (excl.\ Meta-V); Meta-V Avg.\ = unweighted mean of short, medium and long.} \\
                   & Default prompts   & \ecwrap{\echint{One line per dataset; \ecQ{} = question, \ecO{} = answer options.}} \\
                   & Video-MME         & \ecwrap{\ecprompt{Select the best answer to the following multiple-choice question based on the video. Respond with only the letter (A, B, C, or D) of the correct option.} \ecQ{} \ecO{} \ecprompt{Answer with the option's letter from the given choices directly.}} \\
                   & LongVideoBench    & \ecwrap{\ecQ{} \ecO{} \ecprompt{Answer with the option's letter from the given choices directly.}} \\
                   & TempCompass       & \ecwrap{\ecQ{} \ecO{} \ecprompt{Please directly give the best option:}} \\
                   & MMVU              & \ecwrap{\ecalt{MC} \ecprompt{Question:}\ecQ{} \ecO{} \ecprompt{Visual Information: processed video Do not generate any intermediate reasoning process. Answer directly with the option letter from the given choices.}\\
                                          \ecalt{open-ended} \ecprompt{Question:}\ecQ{} \ecprompt{Visual Information: processed video Do not generate any intermediate reasoning process. Directly output the final answer.}} \\
                   & LVBench           & \ecwrap{\ecQ{} \ecO{} \ecprompt{Answer the question with the option letter}} \\
                   & VSI-Bench         & \ecwrap{\ecalt{MC} \ecprompt{These are frames of a video.} \ecQ{} \ecprompt{Options:} \ecO{} \ecprompt{Answer with the option's letter from the given choices directly.}\\
                                          \ecalt{numerical} \ecprompt{These are frames of a video.} \ecQ{} \ecprompt{Please answer the question using a single word or phrase.}} \\
                   & MLVU              & \ecwrap{\ecQ{} \ecO{} \ecprompt{Only give the best option. Best option:}} \\
\ecbottomrule
\end{tabular}
\caption{Evaluation card for Table~\ref{tab:table4_reported_vs_ours}. Options in \textbf{bold} are the ones used; a set $\{\cdot\}$ denotes a parameter sweep (one run per value).}
\label{tab:eval-card-table4}
\end{table*}
%
%
\makeatletter
\@ifpackageloaded{xcolor}{}{%
  \PackageError{evaluation_card}{xcolor is not loaded}%
  {Add \string\usepackage{xcolor} to your preamble.}}
\makeatother

\definecolor{cardblue}{HTML}{1A73E8}   
\definecolor{cardrule}{HTML}{C9D7EE}   

\ifdefined\ecgroupw\else\newlength\ecgroupw\fi
\ifdefined\ecfieldw\else\newlength\ecfieldw\fi
\ifdefined\ecvalw\else\newlength\ecvalw\fi

\providecommand{\ecgroup}[1]{\textcolor{cardblue}{\bfseries\textsc{#1}}}      
\providecommand{\echint}[1]{\leavevmode{\color{black!50}\itshape #1}}         
\providecommand{\ecsel}[1]{\textbf{#1}}                                        
\providecommand{\ecsep}{\unskip\ \textup{\textbar}\ \ignorespaces}            
\providecommand{\ecsweep}[1]{$\{$#1$\}$}                                       
\providecommand{\ecwrap}[1]{\raggedright #1\let\\\tabularnewline}                           

\providecommand{\ectitle}[1]{%
  \multicolumn{3}{@{}l@{}}{\large\bfseries\color{cardblue}#1}\\
  \noalign{\vskip 2pt{\color{cardblue}\hrule height 0.8pt}\vskip 4pt}}
\providecommand{\ecbottomrule}{%
  \noalign{\vskip 3pt{\color{cardblue}\hrule height 0.8pt}}}
\providecommand{\ecgroupsep}{%
  \noalign{\vskip 3pt%
    \moveright\dimexpr\ecgroupw+2\tabcolsep\relax
    \vbox{\color{cardrule}\hrule width \dimexpr\linewidth-\ecgroupw-2\tabcolsep\relax height 0.4pt}%
    \vskip 3pt}}

\providecommand{\ecsetwidths}[2]{%
  \settowidth\ecgroupw{\ecgroup{#1}}%
  \settowidth\ecfieldw{#2}%
  \setlength\ecvalw{\dimexpr\linewidth-\ecgroupw-\ecfieldw-4\tabcolsep\relax}}

\providecommand{\ectodo}[1][to fill in]{\textcolor{red!70!black}{\itshape[#1]}}  
\providecommand{\na}{--}
\providecommand{\ecprompt}[1]{{\footnotesize\ttfamily\frenchspacing #1}}      
\providecommand{\ecQ}{\echint{\textrm{$\langle$Q$\rangle$}}}                  
\providecommand{\ecO}{\echint{\textrm{$\langle$O$\rangle$}}}                  
\providecommand{\ecalt}[1]{\echint{\textrm{#1:}}}                             

\begin{table*}[t]
\centering
\small
\renewcommand{\arraystretch}{1.15}
\ecsetwidths{Comments}{Inference backend}
\begin{tabular}{@{}p{\ecgroupw}p{\ecfieldw}p{\ecvalw}@{}}
\ectitle{Evaluation Card \textmd{\normalsize\color{black!60}(Table~\ref{tab:mvb-prompt-modality-ablation})}}
\ecgroup{Scope}    & Models            & Qwen2.5-VL-7B \\
                   & Datasets          & Meta-V \echint{(short / medium / long)} \\
\ecgroupsep
\ecgroup{Runtime}  & Inference backend & HF \\
                   & Batch size        & 1 \echint{(.mp4; .jpg with vt)}; 16 \echint{(.jpg with tvt, vtv)} \\
\ecgroupsep
\ecgroup{Input}    & Input modality    & \ecsweep{.mp4, .jpg} \\
                   & No.\ of frames    & 4 \\
                   & Frame sampling    & \ecsel{uniform}\echint{\ecsep fps rate} \\
                   & Max frames        & \ecsel{none}\echint{\ecsep int} \\
                   & Cap mode          & \ecsel{none}\echint{\ecsep uniform\ecsep cutoff} \\
\ecgroupsep
\ecgroup{Decoding} & Video decoder     & torchcodec \echint{(.mp4 input; n/a for .jpg)} \\
                   & Frame resolution  & 602112 \echint{(max pixel count)} \\
                   & Temperature       & 0.0 \\
                   & Max new tokens    & default\echint{; 1024 for $P1$} \\
                   & Prompt layout     & \ecsweep{vt, tvt, vtv} \echint{(.jpg only; n/a for .mp4)} \\
                   & Prompt(s)         & \ecsweep{$P0$, $P1$, $P^*$} \echint{(see Comments)} \\
\ecgroupsep
\ecgroup{Comments} & Protocol          & \ecwrap{Standardized configuration at 4 frames comparing the input file type: the video passed as a single .mp4 file vs.\ the same uniformly sampled frames passed as .jpg images. Each prompt is run on .mp4 and on .jpg with each of the three layouts. All other settings are held fixed.} \\
                   & Layouts           & \ecwrap{Order of video (v) and text (t) in the prompt: vt = video then text; tvt = text before and after the video; vtv = video before and after the text.} \\
                   & $P0$ (default)    & \ecwrap{Default per-dataset prompt from \texttt{lmms\_eval}.} \\
                   & $P1$ (CoT)        & \ecwrap{\ecprompt{Think step by step about what happens in the video, then end your reply with Answer:} \echint{(max.\ 1024 new tokens)}} \\
                   & $P^*$ (ours)      & \ecwrap{Replaces the default prompt with \ecprompt{Respond with a single letter and nothing else.}} \\
                   & Aggregates        & \ecwrap{Avg.\ = weighted mean of the short, medium and long splits; subscripts give the drop w.r.t.\ $P^*$ in the same input modality and layout.} \\
\ecbottomrule
\end{tabular}
\caption{Evaluation card for Table~\ref{tab:mvb-prompt-modality-ablation}. Options in \textbf{bold} are the ones used; a set $\{\cdot\}$ denotes a parameter sweep (one run per value).}
\label{tab:eval-card-prompt-modality-ablation}
\end{table*}

\begin{table*}[t]
\centering
\small
\renewcommand{\arraystretch}{1.15}
\ecsetwidths{Comments}{Inference backend}
\begin{tabular}{@{}p{\ecgroupw}p{\ecfieldw}p{\ecvalw}@{}}
\ectitle{Evaluation Card \textmd{\normalsize\color{black!60}(Table~\ref{tab:mvb-prompt-layout-ablation})}}
\ecgroup{Scope}    & Models            & Qwen2.5-VL-7B \\
                   & Datasets          & Meta-V \echint{(short / medium / long)} \\
\ecgroupsep
\ecgroup{Runtime}  & Inference backend & HF \\
                   & Batch size        & 16 \\
\ecgroupsep
\ecgroup{Input}    & Input modality    & .jpg \echint{(frames passed as images)} \\
                   & No.\ of frames    & \ecsweep{32, 64, 128} \\
                   & Frame sampling    & \ecsel{uniform}\echint{\ecsep fps rate} \\
                   & Max frames        & \ecsel{none}\echint{\ecsep int} \\
                   & Cap mode          & \ecsel{none}\echint{\ecsep uniform\ecsep cutoff} \\
\ecgroupsep
\ecgroup{Decoding} & Video decoder     & \na \echint{(n/a for .jpg input)} \\
                   & Frame resolution  & 602112 \echint{(max pixel count)} \\
                   & Temperature       & 0.0 \\
                   & Max new tokens    & default\echint{; 1024 for $P1$} \\
                   & Prompt layout     & \ecsweep{vt, tvt, vtv} \\
                   & Prompt(s)         & \ecsweep{$P0$, $P1$, $P^*$} \echint{(see Comments)} \\
\ecgroupsep
\ecgroup{Comments} & Protocol          & \ecwrap{Standardized configuration with a full sweep over number of frames (32, 64, 128), prompt layout (vt, tvt, vtv) and prompt ($P0$, $P1$, $P^*$), i.e.\ 27 runs, all with .jpg input. All other settings are held fixed.} \\
                   & Layouts           & \ecwrap{Order of video (v) and text (t) in the prompt: vt = video then text; tvt = text before and after the video; vtv = video before and after the text.} \\
                   & $P0$ (default)    & \ecwrap{Default per-dataset prompt from \texttt{lmms\_eval}.} \\
                   & $P1$ (CoT)        & \ecwrap{\ecprompt{Think step by step about what happens in the video, then end your reply with Answer:} \echint{(max.\ 1024 new tokens)}} \\
                   & $P^*$ (ours)      & \ecwrap{Replaces the default prompt with \ecprompt{Respond with a single letter and nothing else.}} \\
                   & Aggregates        & \ecwrap{Avg.\ = weighted mean of the short, medium and long splits; subscripts give the difference w.r.t.\ $P^*$ at the same number of frames and layout ($\downarrow$ lower, $\uparrow$ higher).} \\
\ecbottomrule
\end{tabular}
\caption{Evaluation card for Table~\ref{tab:mvb-prompt-layout-ablation}. Options in \textbf{bold} are the ones used; a set $\{\cdot\}$ denotes a parameter sweep (one run per value).}
\label{tab:eval-card-prompt-layout-ablation}
\end{table*}

\subsection{Tiling in InternVL models}
\begin{table*}[t]
\centering
\caption{InternVL3.5 accuracy (\%) versus number of input frames and maximum number of tiles per frame (\texttt{max\_num\_patches}; second header row). Input: raw \texttt{.mp4} decoded to exactly $N$ frames, tiled video path (\texttt{hf\_tiled}). HF backend, fixed seed.
Bold: best setting per row. 
}
\setlength{\tabcolsep}{3pt}
\resizebox{\textwidth}{!}{%
\begin{tabular}{ll|ccccc|ccccc|cccc|cc|c|c}
\toprule
 & & \multicolumn{5}{c}{4 frames} & \multicolumn{5}{c}{8 frames} & \multicolumn{4}{c}{16 frames} & \multicolumn{2}{c}{32 frames} & \multicolumn{1}{c}{64 frames} & \multicolumn{1}{c}{128 frames} \\
\cmidrule(lr){3-7}\cmidrule(lr){8-12}\cmidrule(lr){13-16}\cmidrule(lr){17-18}\cmidrule(lr){19-19}\cmidrule(lr){20-20}
Model & Dataset & 1 & 2 & 4 & 8 & 12 & 1 & 2 & 4 & 8 & 12 & 1 & 2 & 4 & 8 & 1 & 2 & 1 & 1 \\
\midrule
\multirow{7}{*}{InternVL3.5-4B} & Video-MME & 54.6 & 53.4 & 53.4 & 53.2 & 53.2 & 57.7 & 58.1 & 58.2 & 56.5 & 56.5 & 60.7 & 60.8 & 60.8 & 58.7 & 63.1 & 63.0 & \textbf{64.8} & 64.1 \\
 & MVBench & 67.0 & 67.9 & 67.8 & 66.5 & 66.7 & 70.0 & 70.6 & 70.5 & 68.2 & 68.0 & 71.4 & 71.7 & 71.8 & 67.1 & \textbf{71.9} & 71.5 & 71.3 & 69.4 \\
 & TempCompass (MC) & 69.2 & 68.2 & 68.2 & 68.2 & 68.2 & 68.9 & 68.8 & 68.8 & 68.8 & 68.8 & 71.1 & 70.2 & 70.2 & 70.2 & \textbf{71.4} & 69.4 & 69.9 & 68.9 \\
 & MLVU (dev) & 59.6 & 59.7 & 59.5 & 58.2 & 57.0 & 60.6 & 61.1 & 61.1 & 60.8 & 60.1 & 63.4 & 63.3 & 62.8 & 62.6 & 67.1 & 67.2 & 67.7 & \textbf{68.5} \\
 & LongVideoBench (val) & 52.1 & 52.7 & 52.6 & 52.3 & 52.3 & 54.5 & 54.4 & 54.4 & 54.3 & 53.9 & 56.8 & 57.6 & 57.6 & 55.9 & 58.3 & 57.8 & \textbf{59.2} & 58.9 \\
 & LVBench & 38.0 & 37.4 & 37.5 & 39.3 & 39.1 & 38.9 & 38.2 & 38.2 & 38.3 & 38.0 & 41.8 & 40.7 & 40.9 & 40.4 & 42.7 & 41.0 & 43.2 & \textbf{44.7} \\
 & MMVU (val) & 46.4 & 47.1 & 47.1 & 47.1 & 47.3 & 46.7 & 48.9 & 48.9 & 47.0 & 47.7 & 48.6 & 49.2 & 49.2 & 49.1 & \textbf{50.8} & 49.0 & 49.9 & 49.3 \\
\midrule
\multirow{7}{*}{InternVL3.5-8B} & Video-MME & 56.8 & 55.7 & 55.7 & 55.5 & 55.5 & 60.1 & 58.7 & 58.7 & 58.7 & 58.7 & 63.0 & 61.8 & 61.8 & 59.8 & 64.8 & 63.3 & \textbf{66.3} & 65.2 \\
 & MVBench & 68.4 & 68.4 & 68.4 & 66.8 & 66.6 & 72.0 & 71.7 & 71.7 & 69.2 & 69.0 & 73.3 & 72.5 & 72.6 & 68.2 & \textbf{73.4} & 72.0 & 72.7 & 71.1 \\
 & TempCompass (MC) & 69.4 & 69.4 & 69.4 & 69.4 & 69.4 & 70.3 & 69.4 & 69.4 & 69.4 & 69.4 & 70.6 & 69.6 & 69.6 & 69.6 & \textbf{70.8} & 69.4 & 70.5 & 70.1 \\
 & MLVU (dev) & 58.2 & 58.3 & 58.0 & 57.9 & 57.1 & 62.5 & 62.3 & 62.1 & 60.6 & 59.6 & 63.8 & 63.3 & 63.5 & 62.1 & 67.8 & 67.3 & \textbf{68.5} & \textbf{68.5} \\
 & LongVideoBench (val) & 55.0 & 53.9 & 53.8 & 54.4 & 54.5 & 56.5 & 56.4 & 56.3 & 56.1 & 55.8 & 57.6 & 58.8 & 59.0 & 56.3 & 59.9 & 59.8 & \textbf{62.9} & 61.6 \\
 & LVBench & 38.7 & 39.7 & 39.6 & 39.6 & 39.4 & 39.6 & 39.4 & 39.4 & 40.6 & 40.5 & 40.6 & 42.0 & 42.2 & 41.2 & 41.9 & 41.6 & 43.9 & \textbf{46.2} \\
 & MMVU (val) & 55.7 & 56.8 & 56.8 & 55.7 & 57.0 & 55.4 & 56.8 & 56.8 & 57.8 & 56.3 & 57.7 & 59.1 & 59.1 & 58.3 & 56.4 & 59.6 & \textbf{59.9} & 58.3 \\
\bottomrule
\end{tabular}}

\label{tab:internvl35_tiles_exp5}
\end{table*}

\begin{table*}[t]
\centering
\caption{InternVL3.5 accuracy (\%) versus number of input frames and maximum number of tiles per frame (\texttt{max\_num\_patches}; second header row). Input: $N$ pre-extracted \texttt{.jpg} frames passed as images, stock HF image tiler (\texttt{hf}). HF backend, fixed seed.
Bold: best setting per row. 
}
\setlength{\tabcolsep}{3pt}
\resizebox{\textwidth}{!}{%
\begin{tabular}{ll|ccccc|ccccc|cccc|cc|c|c}
\toprule
 & & \multicolumn{5}{c}{4 frames} & \multicolumn{5}{c}{8 frames} & \multicolumn{4}{c}{16 frames} & \multicolumn{2}{c}{32 frames} & \multicolumn{1}{c}{64 frames} & \multicolumn{1}{c}{128 frames} \\
\cmidrule(lr){3-7}\cmidrule(lr){8-12}\cmidrule(lr){13-16}\cmidrule(lr){17-18}\cmidrule(lr){19-19}\cmidrule(lr){20-20}
Model & Dataset & 1 & 2 & 4 & 8 & 12 & 1 & 2 & 4 & 8 & 12 & 1 & 2 & 4 & 8 & 1 & 2 & 1 & 1 \\
\midrule
InternVL3.5-4B & TempCompass (MC) & 64.6 & 65.4 & 65.4 & 65.4 & 65.4 & 66.8 & 67.0 & 67.0 & 67.0 & 67.0 & \textbf{68.4} & 66.6 & 66.6 & 66.6 & \textbf{68.4} & 66.3 & \textbf{68.4} & 66.1 \\
 & MMVU (val) & 46.6 & 46.4 & 46.4 & 47.6 & 44.0 & 46.1 & 47.1 & 47.1 & 47.1 & 45.1 & 47.1 & 47.9 & 47.9 & \textbf{48.2} & \textbf{48.2} & 47.9 & 47.8 & 47.5 \\
\midrule
InternVL3.5-8B  & TempCompass (MC) & 65.0 & 65.3 & 65.3 & 65.3 & 65.3 & 66.8 & 66.8 & 66.8 & 66.8 & 66.8 & 68.2 & 67.6 & 67.6 & 67.6 & \textbf{69.1} & 67.7 & 67.8 & 66.8 \\
 & MMVU (val) & 54.1 & \textbf{57.4} & \textbf{57.4} & 55.1 & 54.4 & 55.6 & 56.3 & 56.3 & 56.3 & 55.7 & 56.4 & 55.8 & 55.8 & 55.6 & 55.4 & 56.1 & 54.6 & 55.1 \\
\bottomrule
\end{tabular}}

\label{tab:internvl35_tiles_exp7}
\end{table*}

In Tables~\ref{tab:internvl35_tiles_exp5} and~\ref{tab:internvl35_tiles_exp7}, 
we show the results for InternVL3.5-4b and InternVL3.5-8b for multiple datasets with different maximum numbers of tiles splitting each frame into up to k high-resolution tiles \texttt{max\_num\_patches} used per frame. Using a higher number of tiles per frame only helps at a lower number of frames for a few datasets. For example, for MMVU dataset InternVL3.5-4b (with 4 frames) model obtains 0.9 higher performance with 12 tiles compared with one tile. Similar trends can be observed for the InternVL3.5-8b model. In Table~\ref{tab:internvl35_tiles_exp7}, we show results for InternVL3.5 models with .jpg as input. A similar trend can be observed, where having more tiles per frame helps at lower frame counts, such as 4 and 8 frames.

\end{document}